\pdfoutput=1 

\newif\REVIEW

\ifdefined\REVIEW
    \documentclass[
        a4paper,
    ]{article}
    \usepackage{lineno, setspace}
\else
    \documentclass[
        a4paper,
        twocolumn
    ]{article}
\fi

\usepackage{IEK10} \usepackage{natbib}

\def\IEK10{
  Forschungszentrum Jülich GmbH,
  Institute of Climate and Energy Systems,
  Energy Systems Engineering (ICE-1),
  Jülich 52425,
  Germany
}
\def\SVT{
  RWTH Aachen University,
  Process Systems Engineering (AVT.SVT),
  Aachen 52074,
  Germany
}
\def\JARA{
  JARA-ENERGY,
  Jülich 52425,
  Germany
}
\def\RWTH{
  RWTH Aachen University,
  Aachen 52062,
  Germany
}

\newcommand{\mytitle}{
Sample-Efficient Reinforcement Learning of Koopman eNMPC
}

\newcommand{\affil}{
  \begin{itemize}[leftmargin=3mm, itemsep=0mm]
    \item[$^a$]\IEK10
    \item[$^b$]\RWTH
    \item[$^c$]\JARA
    \item[$^d$]\SVT
  \end{itemize}
}

\def\firstAuthor{Daniel Mayfrank}
\newcommand{\myauthor}{
\firstAuthor$^{a,b}$\orcidlink{0009-0000-6275-0614},
Mehmet Velioglu$^{a,b}$\orcidlink{0009-0002-1144-0661},
Alexander Mitsos$^{c,a,d}$\orcidlink{0000-0003-0335-6566}, 
Manuel Dahmen$^{a,*}$\orcidlink{0000-0003-2757-5253} }
\newcommand{\correspondingauthor}{Manuel Dahmen, \IEK10 \newline E-mail: m.dahmen@fz-juelich.de}
\author{\myauthor}

\usepackage[
  colorlinks,
  linkcolor=blue,
  citecolor=blue,
  urlcolor=blue,
  pdftitle={\mytitle},
  pdfauthor={\firstAuthor}
]{hyperref}
\usepackage[capitalise, nameinlink]{cleveref}
\crefname{table}{Tab.}{Tab.}

\usepackage{orcidlink}

\newcommand{\setpgfexternalcounter}[1]{
  \makeatletter \pgfkeysgetvalue{/tikz/external/figure name}\myexternalname
  \expandafter\gdef\csname c@tikzext@no@\myexternalname\endcsname{#1}\makeatother
}

\begin{document}

\ifx\REVIEW\undefined
\twocolumn[
\begin{@twocolumnfalse}
\fi
  \thispagestyle{firststyle}

  \begin{center}
    \begin{large}
      \textbf{\mytitle}
    \end{large} \\
    \myauthor
  \end{center}

  \vspace{0.5cm}

  \begin{footnotesize}
    \affil
  \end{footnotesize}

  \vspace{0.5cm}

    Reinforcement learning (RL) can be used to tune data-driven (economic) nonlinear model predictive controllers ((e)NMPCs) for optimal performance in a specific control task by optimizing the dynamic model or parameters in the policy’s objective function or constraints, such as state bounds.
    However, the sample efficiency of RL is crucial, and to improve it, we combine a model-based RL algorithm with our published method that turns Koopman (e)NMPCs into automatically differentiable policies.
    We apply our approach to an eNMPC case study of a continuous stirred-tank reactor (CSTR) model from the literature. The approach outperforms benchmark methods, i.e., data-driven eNMPCs using models based on system identification without further RL tuning of the resulting policy, and neural network controllers trained with model-based RL, by achieving superior control performance and higher sample efficiency.
    Furthermore, utilizing partial prior knowledge about the system dynamics via physics-informed learning further increases sample efficiency.

  \vspace*{5mm}
\ifx\REVIEW\undefined
\end{@twocolumnfalse}
]
\fi

\newpage

\section{Introduction}\label{sec:intro}
    Model predictive control (MPC) and its variants, e.g., economic nonlinear MPC (eNMPC), rely on dynamic models that (i) are accurate and (ii) lead to optimal control problems (OCPs) which are solvable in real-time. In process systems engineering, obtaining mechanistic models that fulfill these requirements can be difficult due to large scales, unknown parameters, and nonlinearities (\cite{tang2022data}). As an alternative to mechanistic models, data-driven approaches can be employed. These approaches are typically based on system identification (SI) using either historical operating data or data generated with a mechanistic model of the physical system (\cite{tang2022data}).

\begin{figure}[h]
        \centering
        \subfloat[a][]{\includegraphics[width=0.3\paperwidth]{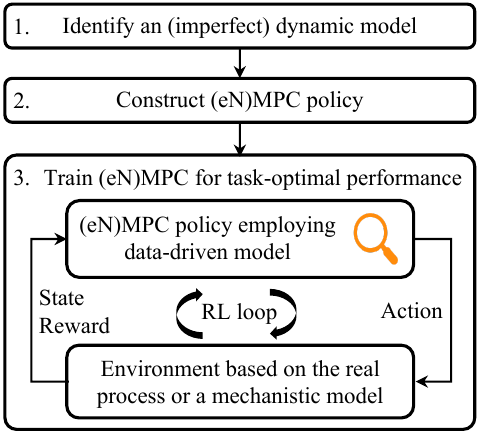} \label{fig:E2E}} \\
        \subfloat[b][]{\includegraphics[width=0.2\paperwidth]{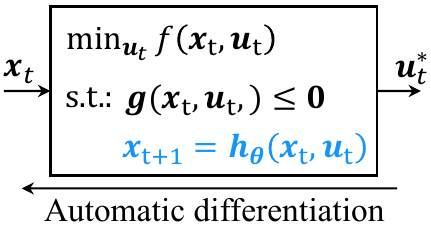} \label{fig:E2E_model}}
        \subfloat[c][]{\includegraphics[width=0.2\paperwidth]{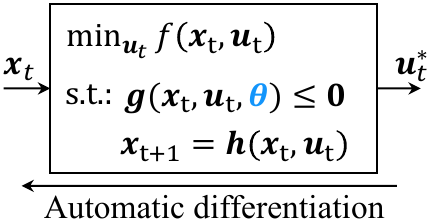} \label{fig:E2E_bounds}}
        \caption{(a) Procedure for RL-based training of an (eN)MPC. (b) A differentiable eNMPC policy parameterized by the parameters $\bm{\theta}$ of the dynamic model; takes as input the current state $\bm{x}_t$ and computes the optimal control action $\bm{u}^{*}_{t}$ based on the minimization of a cost function $f$, subject to inequality constraints $\bm{g}$, and the learnable dynamic model $\bm{h_\theta}$. (c) Differentiable eNMPC policy with parameterized inequality constraints $\bm{g}$, e.g., state bounds. Training this policy leaves the underlying dynamic model $\bm{h}$ unchanged but adapts the inequality constraints to counteract model-plant mismatch.} \label{fig:E2E_intro}
    \end{figure}

    Alternatively, using reinforcement learning (RL) methods, data-driven (eN)MPCs can be trained for optimal performance in specific control tasks (see Fig. \ref{fig:E2E}), which may produce superior control performance compared to SI (see, e.g., \cite{chen2019gnu, gros2019data, mayfrank2024KoopmanPPO, mayfrank2024KoopmanSHAC}).
    To this end, a differentiable (e)NMPC policy is constructed, wherein the learnable parameters can be parameters of the dynamic model (see Fig. \ref{fig:E2E_model}, e.g., \cite{mayfrank2024KoopmanPPO, mayfrank2024KoopmanSHAC}), or other parameters that appear in the objective function or constraints of the (e)NMPC (\cite{gros2019data, brandner2023reinforcement, brandner2024reinforced}), e.g., the state bounds (see Fig. \ref{fig:E2E_bounds}). The latter approach optimizes the policy by compensating for model errors, e.g., via bounds adaptation. Thereby, RL-based training/refinement of a highly parameterized dynamic model such as an artificial neural network can be avoided by optimizing few (interpretable) parameters, e.g., parameters that modify the state bounds. Therefore, this approach may lead to better convergence, compared to RL-based (re)training of the dynamic model itself. However, even positing good training convergence, the model-free\footnote{Due to the multitude of ways in which models can be used in RL, the distinction between model-free RL, model-based RL, and other control approaches is not consistent across the literature. This distinction might be especially confusing in the context of the present work since we study the training of eNMPC policies, i.e., inherently model-based policies. However, the distinction between model-free and model-based, which is most relevant to our work, is about whether the \textit{training algorithm} uses an additional learned model of the environment to train the policy. Please refer to Section \ref{sec:method_mbpo} for the exact definitions that we use in this work.}
    RL algorithms that have so far been used for RL-based training of (eN)MPCs remain notoriously sample inefficient, essentially rendering them inapplicable to domains where interacting with the physical environment (sampling) is expensive (\cite{gopaluni2020modern}), e.g., in industrial chemical process control applications.

    Model-based RL (MBRL) algorithms are designed to increase the sample efficiency of RL by concurrently learning a policy and a model of the environment, i.e., a function that allows the prediction of future states and rewards from state-action pairs.
    \begin{figure}[ht]
    	\centering
    	\includegraphics[width=0.32\paperwidth]{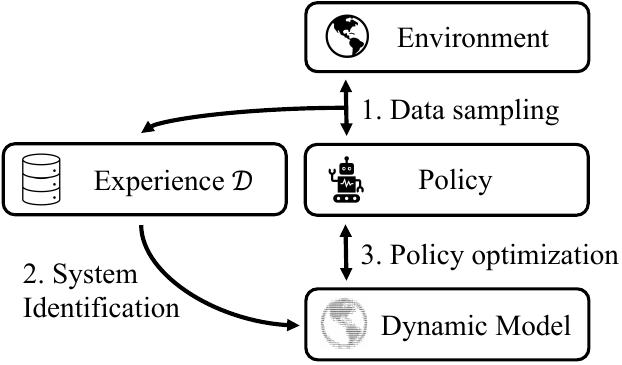}
    	\caption{Dyna-style (\cite{sutton1991dyna}) model-based RL framework. The three steps are repeated for a predefined number of steps, or until satisfactory control performance is reached.}
    	\label{fig:dyna-style_mbrl}
    \end{figure}
    The seminal Dyna algorithm (\cite{sutton1991dyna}) iterates between three steps (see Figure \ref{fig:dyna-style_mbrl}): (i) The current policy interacts with the real environment to gather data about the system dynamics. (ii) Using the acquired data, a data-driven dynamic model of the environment is learned via SI. (iii) The learned model is used to optimize the policy via any suitable RL algorithm, e.g., Proximal Policy Optimization (PPO) (\cite{schulman2017proximal}) or Soft Actor-Critic (\cite{haarnoja2018soft}). Numerous ``Dyna-style algorithms'', i.e., algorithms that follow this basic framework, have been developed over the years. However, in this framework, the policy can learn to exploit the errors of the dynamic model, leading to overly-optimistic simulated results and corresponding policy failures in the real world (\cite{kurutach2018model}). However, recent contributions (\cite{kurutach2018model, clavera2018model, Janner2019MBPO}) have showcased ways to counteract this problem based on learning ensembles of dynamic models, leading to algorithms that can match the asymptotic control performance of model-free RL on certain problems while requiring orders of magnitude fewer interactions with the real environment. Ultimately, to maximize sample efficiency, any Dyna-style algorithm relies on learning reasonably well-generalizing dynamic models of the environment with as little data as possible. Therefore, another intuitive way to improve the performance of Dyna-style algorithms is to incorporate prior knowledge of the dynamics of the environment into the models, e.g., using physics-informed learning. Multiple contributions (e.g., \cite{liu2021PIMBRL, Ramesh2023PIMBRL}) have shown that using physics-informed neural networks (PINNs) (\cite{raissi2019physics}) in Dyna-style algorithms can increase both the sample efficiency and the performance of the resulting policies.

    The adoption of RL algorithms into the process control community is still in its infancy and has largely been limited to model-free RL (\cite{faria2022reinforcement, Dogru_RLProcess2024}). \cite{gopaluni2020modern} describe model-free RL algorithms as insufficiently data-efficient for industrial process control applications. They identify the unification of model-based and model-free methods as a research area with tremendous potential to redefine automation in the process industry. \cite{ponse2024reinforcement} provide a comprehensive review on the applications of RL in the field of sustainable energy systems control, which is adjacent to process control and rate model-based methods as ``underexplored''. Contributions that did leverage model-based RL methods for process control or energy systems control used neural network policies that map directly from states to actions (e.g., \cite{zhang2021robust, gao2023comparative, faridi2024advancing}). On the other hand, numerous contributions have showcased the potential benefits of RL-based training of (eN)MPCs for control compared to SI (e.g., \cite{chen2019gnu, gros2019data, mayfrank2024KoopmanPPO, mayfrank2024KoopmanSHAC}). However, these contributions used model-free RL algorithms. Due to the substantial benefits of model-based RL algorithms compared to model-free variants regarding sample efficiency, it is tempting to examine whether model-based RL can accelerate RL-based (eN)MPC training, thus making it potentially feasible in a wider range of applications. Such an analysis has not been conducted thus far.

    In our previous publication (\cite{mayfrank2024KoopmanPPO}), we introduced a method for RL training of (e)NMPCs that utilize data-driven Koopman \textit{surrogate} models. That method rests on the availability of a simulated RL environment based on an accurate mechanistic system model. When training in a simulated environment, sample efficiency is less critical than in many potential real-world RL setting, where the agent has to learn from (costly) interactions with the physical system. In the current work, we extend the applicability of RL-based training of Koopman (e)NMPCs to settings where sample efficiency is critical by combining our previously published approach with the Model-Based Policy Optimization (MBPO) (\cite{Janner2019MBPO}) algorithm. Moreover, we further increase the sample efficiency by modifying MBPO to utilize partial prior knowledge of the dynamics of the controlled system through physics-informed learning. To our knowledge, this work is the first to connect RL-based training of (eN)MPCs for task-optimal performance in specific control tasks to Dyna-style model-based RL.

    In this work, we choose the MBPO algorithm (\cite{Janner2019MBPO}), since it is a state-of-the-art Dyna-style RL algorithm. Many model-based RL algorithms exist and it is impossible to know a priori which algorithm will perform best in a specific task (\cite{wang2019benchmarking}). While MBPO is one of the most promising algorithms available, our approach should be compatible with any Dyna-style algorithm that does not require a specific policy architecture.

    We test our proposed method on an eNMPC case study (\cite{mayfrank2024KoopmanPPO}) based on a continuous stirred-tank reactor (CSTR) model from \cite{flores2006simultaneous}. We assess the performance of our approach by comparing it to that of (i) Koopman eNMPCs trained iteratively via SI and (ii) neural network policies trained using (physics-informed) MBPO. We find that through the combination of iterative SI of the Koopman model that is utilized in the eNMPC and RL-based adaptation of the state bounds in the eNMPC via the MBPO algorithm, our method outperforms the benchmarks for this case study. Additionally, we find that physics-informed learning of the model ensemble that is used in MBPO offers benefits for sample efficiency and that it can prevent policy degradation during training. These findings confirm our expectation that model-based RL can be successfully integrated with training data-driven (eN)MPCs. Thus, our work is a step toward making RL-based training of predictive controllers feasible for complex real-world control problems where no simulator of the environment is available a priori and interactions with the real environment are expensive, making sample efficiency absolutely crucial. Considering the previously mentioned contributions showcasing the advantages of RL compared to pure SI when learning data-driven predictive controllers, our work, therefore, shows an avenue toward more capable and efficient predictive controllers.

    The remainder of this paper is structured as follows: Section \ref{sec:method} provides the theoretical background to our work and presents our method. Section \ref{sec:results} presents the results of the numerical experiments that we conducted on a simulated case study. Section \ref{sec:conclusion} draws final conclusions and discusses promising directions for future research.

\section{Method}\label{sec:method}
    Subsection \ref{sec:method_mbpo} introduces notation and definitions for policy optimization using RL and provides a brief explanation of the MBPO algorithm (\cite{Janner2019MBPO}). Subsection \ref{sec:method_e2eKoopman} explains how we set up differentiable Koopman (e)NMPCs that are trainable using RL. Subsequently, we present our method for sample-efficient learning of task-optimal Koopman (e)NMPCs for control (Subsection \ref{sec:method_PIMBRL-Koopman}).

\subsection{Model-Based Policy Optimization}\label{sec:method_mbpo}
    RL is a framework for learning how to map situations to actions in order to maximize a numerical reward signal (\cite{sutton2018reinforcement}). In contrast to supervised learning tasks where learning is based on labeled data sets, RL is based on sequential feedback from trial and error actuation of an \textit{environment}. The environment is represented by a Markov Decision Process (MDP) with associated states $\bm{x}_t \in \mathbb{R}^n$, control inputs $\bm{u}_t \in \mathbb{R}^m$, a transition function $\bm{\mathcal{F}}: \mathbb R^n \times \mathbb R^m \to \mathbb R^n$,
    \begin{equation}
        \bm{x}_{t+1} = \bm{\mathcal{F}}(\bm{x}_{t},\bm{u}_{t}), \label{eq:transition_function}
    \end{equation}
    and a scalar reward function $\mathcal{R}: \mathbb R^n \times \mathbb R^m \to \mathbb{R}$,
    \begin{equation}
        r_{t+1} = \mathcal{R}(\bm{x}_{t+1},\bm{u}_{t}). \label{eq:reward_function}
    \end{equation}
    The goal of RL is to maximize the (discounted) sum of expected future rewards. For applications with continuous action spaces, actor-critic RL methods (see, e.g., \cite{fujimoto2018addressing, schulman2017proximal, haarnoja2018soft}) that optimize parameterized policies $\bm{\pi}_{\bm{\theta}}(\bm{u}_{t}|\bm{x}_{t})\colon \mathbb{R}^n \mapsto \mathbb{R}^m$ directly mapping from states to (probability distributions over) actions, are most suitable (\cite{sutton2018reinforcement}).

    The Model-Based Policy Optimization (MBPO) (\cite{Janner2019MBPO}) algorithm is a state-of-the-art Dyna-style RL algorithm. As any other Dyna-style algorithm, it iterates between the following three steps (see Fig. \ref{fig:dyna-style_mbrl}): (i) collect experience in the real environment, (ii) learn a model of the environment, (iii) train the policy using the environment model and a suitable model-free RL algorithm. Building upon the work by \cite{kurutach2018model}, MBPO aims to overcome the critical problem of model exploitation by learning an ensemble of models and choosing one of the models at random for each environment step when training the policy (Fig. \ref{fig:dyna-style_mbrl}, third step). Ideally, learning a model ensemble maintains an adequate level of uncertainty in the policy optimization step, thus preventing overfitting the policy to the errors of a specific model. However, model errors still compound over multiple simulated steps, causing problems in policy learning for long-horizon tasks. In MBPO, \cite{Janner2019MBPO} address this issue by introducing two modifications compared to the standard Dyna framework: (i) Instead of using simulated rollouts with $l$ discrete time steps corresponding to the length of episodes in the real environment, they shorten the simulated rollouts to a length of $k \ll l$ steps. (ii) The initial state of each simulated rollout is determined by randomly sampling from the experience $\mathcal{D}$ (cf. Fig. \ref{fig:dyna-style_mbrl}) instead of sampling from the initial state distribution of the real environment. Thus, every state that was encountered by the policy in the real environment can serve as an initial state in a simulated rollout. These modifications disentangle the length of simulated rollouts during policy training from the episode length in the original task. \cite{Janner2019MBPO} showed that in multiple continuous control benchmark problems, MBPO vastly improves sample efficiency while producing policies of similar performance compared to state-of-the-art model-free RL algorithms (e.g., \cite{schulman2017proximal, haarnoja2018soft}).

\subsection{Differentiable Koopman (e)NMPC}\label{sec:method_e2eKoopman}
    In \cite{mayfrank2024KoopmanPPO}), we introduce a method for constructing automatically differentiable stochastic (e)NMPC policies $\bm{\pi}_{\bm{\theta}}(\bm{u}_{t}|\bm{x}_{t})\colon \mathbb{R}^n \mapsto \mathbb{R}^m$ from Koopman models of the form proposed by \cite{korda2018linear}. Such models are of the form
    \begin{subequations}\label{eq:Koopman_model}
    \begin{align}
        \bm{z}_0 &= \bm{\psi}_{\bm{\theta}}(\bm{x}_0),\\
        \bm{z}_{t+1} &= \bm{A}_{\bm{\theta}} \bm{z}_t + \bm{B}_{\bm{\theta}} \bm{u}_t,\\
        \hat{\bm{x}}_t &= \bm{C}_{\bm{\theta}} \bm{z}_t,
    \end{align}
    \end{subequations}
    where $\bm{z}_t \in \mathbb{R}^N$ is the vector of Koopman states and $\hat{\bm{x}}_t \in \mathbb{R}^n$ is the model prediction of the system state at time step $t$. The model has the following components: $\bm{\psi}_{\bm{\theta}}\colon \mathbb{R}^n \mapsto \mathbb{R}^N$, where typically $N \gg n$, defines the nonlinear state observation function that transforms the initial condition $\bm{x}_0$ into the Koopman space. $\bm{A}_{\bm{\theta}} \in \mathbb{R}^{N \times N}$ and $\bm{B}_{\bm{\theta}} \in \mathbb{R}^{N \times m}$ linearly advance the Koopman state vector forward in time. $\bm{C}_{\bm{\theta}} \in \mathbb{R}^{n \times N}$ linearly maps a prediction of the Koopman state to a prediction of the system state.

    Given a data set describing the dynamics of some system, such models can be trained via SI by minimizing the sum of three loss functions (\cite{lusch2018deep, mayfrank2024KoopmanPPO}). These loss terms correspond to the requirements that the Koopman model needs to fulfill: (i) reconstructing states passed through the autoencoder, (ii) predicting the evolution of the lifted Koopman state, and (iii) predicting the evolution of the system states. The associated loss terms are:
    \begin{subequations}
    \label{eq:koopman_SI_losses}
    \begin{align}
        || \bm{C_{\theta}} \bm{\psi_{\theta}}(\bm{x}_t) - \bm{x}_t ||, \label{eq:koopman_SI_losses_AE}\\
        || \bm{A}_{\bm{\theta}} \bm{\psi_{\theta}}(\bm{x}_t) + \bm{B}_{\bm{\theta}} \bm{u}_t - \bm{\psi_{\theta}}(\bm{x}_{t+1}) ||, \label{eq:koopman_SI_losses_pred}\\
        || \bm{C_{\theta}} (\bm{A}_{\bm{\theta}} \bm{\psi_{\theta}}(\bm{x}_t) + \bm{B}_{\bm{\theta}} \bm{u}_t) - \bm{x}_{t+1} || \label{eq:koopman_SI_losses_comb}
    \end{align}
    \end{subequations}

    In \cite{mayfrank2024KoopmanPPO}, we aim to optimize a Koopman model for optimal performance as part of an (e)NMPC in a specific control task. However, as noted in Section \ref{sec:intro}, in RL-based training of an (eN)MPC, it may be beneficial to keep an imperfect model unchanged and instead optimize a small number of parameters in the objective function or inequality constraints which compensate for model errors. To be able to differentiate between different kinds of learnable parameters of the Koopman-eNMPC policy, we therefore rename the parameters: In the following, $\bm{\theta}_\text{K}$ refers to the parameters of the Koopman model, which appear as $\bm{\theta}$ in Eq. \eqref{eq:Koopman_model}. Additionally, we introduce the parameters $\bm{\theta}_{\text{B}}$, which modify the state bounds of the eNMPC. Both types of parameters influence the behavior of the policy, i.e., $\bm{\theta} = [\bm{\theta}^{\intercal}_\text{K}, \bm{\theta}^{\intercal}_{\text{B}}]^{\intercal}$ and $\bm{\pi}_{\bm{\theta}}(\bm{u}_{t}|\bm{x}_{t})\colon \mathbb{R}^n \mapsto \mathbb{R}^m$.

    Integrating the idea of state bound adaptation into the differentiable Koopman (e)NMPC framework (\cite{mayfrank2024KoopmanPPO}) is straightforward. Given an (e)NMPC horizon of $t_{f}+1$ steps with the corresponding sets $\mathrm{T}_{+1} = \{ t,\dots,t+t_f \}$ and $\mathrm{T} = \{ t,\dots,t+t_{f}-1 \}$, a convex OCP is solved to obtain the optimal action $\bm{u}^*_t$:
    \begin{subequations}\label{eq:Koopman_OCP_v2}
    	\begin{align}
    		\underset{(\bm{u}_t)_{t \in \mathrm{T}}}{\min} &\sum_{t \in \mathrm{T}_{+1}} \Phi(\bm{C_{{\theta}_\text{K}}}\bm{z}_t,\bm{u}_t) + M\bm{s}^{\intercal}_{t}\bm{s}_t, \label{eq:OCP_obj_func_v2}\\
    		\text{s.t. }&\bm{z}_{t+1} = \bm{A_{{\theta}_\text{K}}} \bm{z}_t + \bm{B_{{\theta}_\text{K}}} \bm{u}_t \quad\forall t \in \mathrm{T}, \label{eq:OCP_state_evolution_v2}\\
    		&\bm{g}(\bm{C_{{\theta}_\text{K}}}\bm{z}_t,\bm{u}_t,\bm{s}_t,\bm{\theta}_{\text{B}}) \leq \bm{0} \quad\forall t \in \mathrm{T}_{+1} \label{eq:OCP_ineq_constraints_v2}
    	\end{align}
    \end{subequations}
    $\Phi$ is a convex function representing the stage cost of the objective function, and $\bm{g}$ are convex inequality constraint functions that can also include bounds on control and state variables. In the latter case, slack variables $\bm{s}_t$ are added to the state bounds to ensure the feasibility of the OCPs. The use of slack variables is penalized quadratically with a penalty factor $M$. Using PyTorch (\cite{paszke2017automatic}) and \textit{cvxpylayers} (\cite{Agrawal2019differentiable}), the output $\bm{u}_t$ of the policy is automatically differentiable with respect to $\bm{x}_t$, $\bm{\theta}_{\text{K}}$, and $\bm{\theta}_{\text{B}}$.

\subsection{Physics-informed MBPO of Koopman models for control}\label{sec:method_PIMBRL-Koopman}
    This section describes our general framework for sample-efficient learning of task-optimal Koopman (e)NMPC policies. A more in-depth description of the implementation details of our method when applied to our specific case study is provided in Sec. \ref{sec:results_training}.

\begin{figure}[!ht]
        \centering
        \subfloat[a][]{\includegraphics[width=0.4\paperwidth]{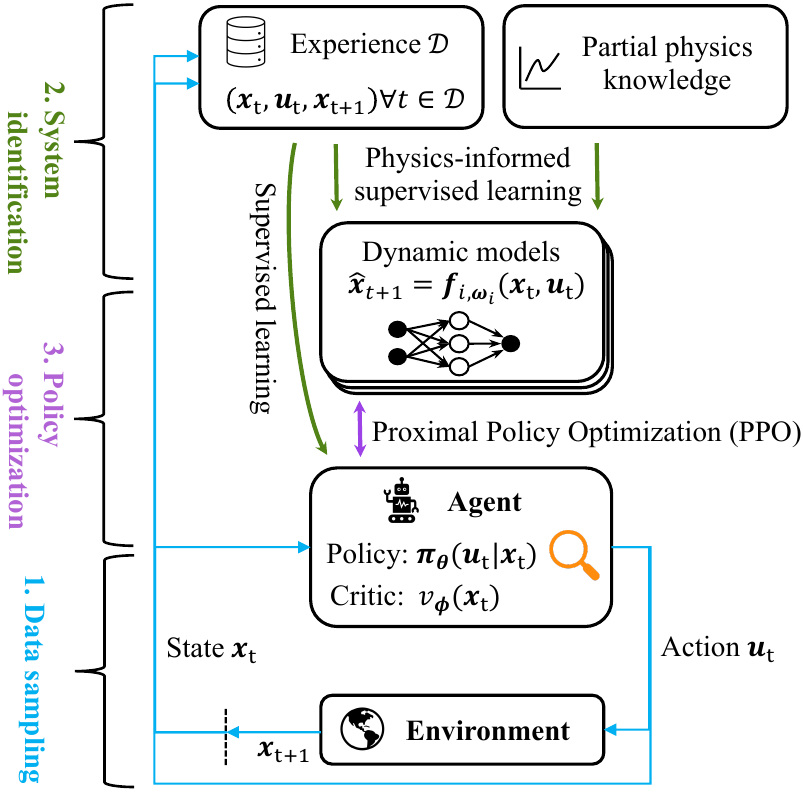} \label{fig:method_mbpo}} \\
        \subfloat[b][]{\includegraphics[width=0.45\paperwidth]{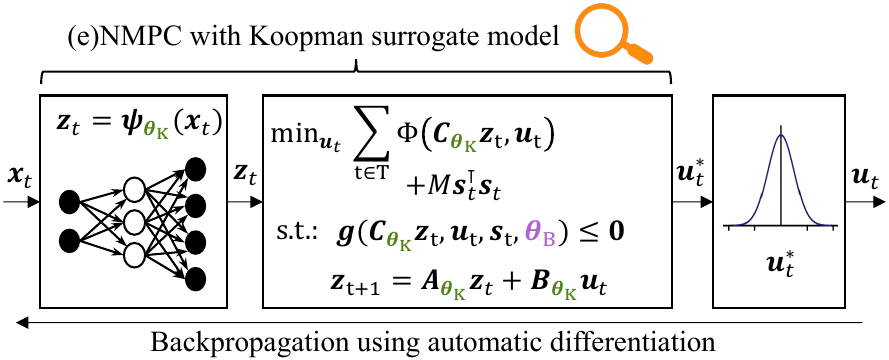} \label{fig:Koopman_eNMPC}}
        \caption{Using MBPO to train a task-optimal Koopman (e)NMPC controller. (a) The training algorithm. The following three steps are executed in a loop until a stopping criterion is reached: First, the Koopman (e)NMPC interacts with the environment to gather data about the dynamics. Second, all data collected up to the current step is used to fit the Koopman model (parameters $\bm{\theta}_{\text{K}}$) and the PINN ensemble (parameters $\bm{\omega}_i \forall i \in \{1,2,\dots,n\}$). Third, a surrogate RL environment is constructed using the NN ensemble and the Koopman (e)NMPC is optimized by tuning the parameters $\bm{\theta}_{\text{B}}$, i.e., the state bounds. (b) The automatically differentiable Koopman (e)NMPC whose behavior is defined by the parameters of the Koopman model ($\bm{\theta}_{\text{K}}$) and the parameters modifying the state bounds ($\bm{\theta}_{\text{B}}$). The parameters are color-coded to match the colors of the corresponding optimization steps in Fig. \ref{fig:method_mbpo}.} \label{fig:method}
    \end{figure}

    Our method iterates between the three typical Dyna steps and is visualized in Figure \ref{fig:method}.
    First, the Koopman (e)NMPC policy interacts with the environment for a predefined number of steps to gather data $\mathcal{D}$ about the system dynamics. To ensure exploration, we sample the (otherwise deterministic) action $\bm{u}_t$ from a normal distribution $\bm{u}_t \sim \bm{\mathcal{N}}(\bm{u}^*_t,\bm{\sigma}^{2})$. We assume that the reward function of the environment is known. Thus, rewards $r_t$ do not need to be recorded in $\mathcal{D}$. In the very first iteration of the overall algorithm the Koopman model is still randomly initialized and the eNMPC outputs will, therefore, not be meaningful. Therefore, in the data sampling step of the first MBPO iteration, we randomly sample the actions $\bm{u}_t$ from a uniform distribution over the action space. Likewise, any other controller type, e.g., a PID controller, may be used in the first MBPO iteration, although the quality of the resulting training data will be influenced by how diverse the control actions produced by the controller are.

    Second, an ensemble of $n$ data-driven dynamic models, i.e., neural networks (NNs), is learned based on $\mathcal{D}$ via SI to approximate the dynamics of the environment (c.f. Eq. \eqref{eq:transition_function}). If (incomplete) physics knowledge is available, it is possible to train PINNs (\cite{Raissi2019Physics-informedEquations}). Throughout this work, each ensemble member $\bm{NN}_{i,\bm{\omega}_i} \forall i \in \{1,2,\dots,n\}$ is a PINN parameterized by $\bm{\omega}_i$. Furthermore, we fit the parameters $\bm{\theta}_\text{K}$ of the Koopman model to $\mathcal{D}$ via SI. Note that we do not use a physics-informed training method for the Koopman model. Physics-informed training of the Koopman model would, in principle, also be possible. However, the representational capacity of the Koopman model is limited because it is used as part of the real-time eNMPC policy and physics-informed training would necessitate using some of that representational capacity to predict outputs that are not necessary for the eNMPC application, i.e., the \textit{a priori} unknown physics terms (see Fig. \ref{fig:pinn-model}). Since the behavior of the overall Koopman eNMPC policy is optimized with respect to the physics-informed NN ensemble anyway (see Fig. \ref{fig:method_mbpo}, step three), and to keep our method as simple as possible, we decided against a physics-informed system identification approach for the Koopman model. In fitting the PINN ensemble and the Koopman model, we follow standard SI practices, such as splitting $\mathcal{D}$ into a training data set and a validation data set used for early stopping and normalizing the inputs and outputs of the models. We refer the reader to \cite{mayfrank2024KoopmanPPO} for a detailed description of how Koopman models in the form proposed by \cite{korda2018linear} (Eq. \eqref{eq:Koopman_model}) can be trained using system identification.

    The third step is based upon our previously published method (\cite{mayfrank2024KoopmanPPO}) on viewing Koopman (e)NMPC policies as automatically differentiable policies (Fig. \ref{fig:Koopman_eNMPC}): Using the ensemble in conjunction with the known reward function as a simulator of the environment, we train the Koopman (e)NMPC for task-optimal performance by optimizing the parameters $\bm{\theta}_\text{B}$ via the Proximal Policy Optimization (PPO) algorithm (\cite{schulman2017proximal}). During policy optimization using PPO, one of the $n$ PINNs is chosen randomly for each step in the simulated environment. The critic is a feedforward neural network parameterized by $\bm{\phi}$. We use the approach described by \cite{kurutach2018model} to determine when to stop the policy optimization and return to the first step (data sampling): In regular intervals, the performance of the policy is evaluated separately on all $n$ learned models. Once the ratio of models under which the policy improves drops below a certain threshold for too many consecutive PPO iterations, we terminate the policy optimization and return to the data sampling step. A more detailed description of the termination criterion for policy optimization is given in Sec. \ref{sec:results_training_RL}. The overall learning process can continue for a predefined number of steps in the real environment or until a satisfactory performance is achieved.

    We emphasize that the behavior of the resulting (e)NMPC is defined by $\bm{\theta}_\text{K}$ and $\bm{\theta}_\text{B}$ and that both parameter types are optimized in each iteration of the overall algorithm (Fig. \ref{fig:method}): In the SI step, $\bm{\theta}_\text{K}$ is trained to maximize the agreement of the Koopman model to $\mathcal{D}$, i.e., all data available at the time. Then, the policy optimization step optimizes $\bm{\theta}_\text{B}$, i.e., the state bounds, using PPO and simulated interactions with the PINN ensemble. Please note that the policy optimization step does not necessarily result in a tightening of the state bounds: Depending on (i) the disagreement between the Koopman model and the PINN ensemble, and (ii) the reward function specified by the user, it may lead to tightened or relaxed bounds. Therefore, if a sensible reward function is specified, the policy optimization step should not be detrimental to the economic performance of the overall policy.

    \paragraph{Intuitive justification for our approach:} The policy optimization step optimizes the Koopman (e)NMPC policy using a surrogate RL environment. This environment is based on an ensemble of PINNs learned via SI using the same data as the Koopman model. In the following, we provide three rationales for why such an approach offers benefits compared to simply learning a Koopman model via SI and using it inside an (e)NMPC policy without further RL-based optimization of the policy: (i) The PINN ensemble captures the epistemic uncertainty of the learned dynamics (\cite{Janner2019MBPO}), i.e., the uncertainty that arises from insufficient amounts of data. Through RL, the (e)NMPC is forced to behave in a way that is robust with respect to every ensemble member. Without RL, a single learned Koopman model would define the (e)NMPC behavior, which increases the risk of policy failures due to epistemic uncertainty. (ii) The representational capacity of the Koopman model that is utilized in the (e)NMPC is limited since the resulting OCPs (Eq. \eqref{eq:Koopman_OCP_v2}) have to be solved in real-time. Thus, given sufficiently complex system dynamics, the Koopman model might not be able to accurately capture the system dynamics everywhere. In contrast, the representational capacity of the PINN ensemble members has a much higher limit since the ensemble models are not used in online optimization. RL tuning of $\bm{\theta}_\text{B}$ can, therefore, help to compensate for the limitations of (e)NMPC that stem from a limited representational capacity of the utilized Koopman model. (iii) Using RL, the (e)NMPC can be tuned to the requirements of a specific control problem. Specifically, by weighting different parts of the reward function of the simulated RL environment, RL offers a way to tune the behavior of the policy, e.g., to prioritize cost savings or constraint satisfaction.

\section{Numerical experiments}\label{sec:results}

\subsection{Case study description}\label{sec:results_casestudy}
    We demonstrate our method on a demand response case study (\cite{mayfrank2024KoopmanPPO}) based on a benchmark continuous stirred-tank reactor (CSTR) model (\cite{flores2006simultaneous, du2015time}). The following case study description is based on \cite{mayfrank2024KoopmanPPO}, where a more detailed explanation can be found. The states $\bm{x}$ of the model are the dimensionless product concentration $c$ and the dimensionless reactor temperature $T$. The control inputs $\bm{u}$ are the production rate $\rho$ $\left[\frac{1}{\text{h}}\right]$ and the coolant flow rate $F$ $\left[\frac{1}{\text{h}}\right]$. Two nonlinear ordinary differential equations define the dynamics of the system:
    \begin{subequations} 
    \label{eq:CSTR_model}
    	\begin{align}
            \dot{c}(t) = &\:(1-c(t)) \dfrac{\rho(t)}{V} - c(t)ke^{-\frac{N}{T(t)}}, \label{eq:CSTR_model_cdot} \\
            \dot{T}(t) = &\:(T_f - T(t))\dfrac{\rho(t)}{V} + c(t)ke^{-\frac{N}{T(t)}} - F(t) \alpha_c (T(t)-T_c) \label{eq:CSTR_model_Tdot}
    	\end{align}
    \end{subequations}
    Table \ref{tab:CSTR_parameters} lists all parameters appearing in Eq. \eqref{eq:CSTR_model}.
    \begin{table}[ht]
    \centering
    \setlength{\extrarowheight}{0.05cm}
        \caption{CSTR model parameters (\cite{flores2006simultaneous, du2015time}). All parameters except the reaction constant $k$ are dimensionless.}
        \label{tab:CSTR_parameters}
        \begin{tabular}{lll}
            \toprule
                                      &    symbol  & value              \\ \hline
            volume                    &      $V$   & $20$               \\
            reaction constant         &      $k$   & $300\frac{1}{\text{h}}$   \\
            activation energy         &      $N$   & $5$                \\
            feed temperature          &      $T_f$ & $0.3947$           \\
            heat transfer coefficient & $\alpha_c$ & $1.95\cdot10^{-4}$   \\
            coolant temperature       &      $T_c$ & $0.3816$             \\
            \bottomrule
        \end{tabular}
    \end{table}

    For the training of the PINN ensemble in the SI step (see Fig. \ref{fig:method_mbpo}), we assume that the concentration and temperature changes due to inlet/outlet flows and cooling are known. However, we assume that the constitutive expression for the reaction in Eqs. \eqref{eq:CSTR_model} is unknown. Thus, the physics equations for the PINN are
    \begin{subequations} 
    \label{eq:PINN_model}
    	\begin{align}
            \dot{c}(t) = &\:(1-c(t)) \dfrac{\rho(t)}{V} - R(t), \label{eq:PINN_model_cdot} \\
            \dot{T}(t) = &\:(T_f - T(t))\dfrac{\rho(t)}{V} + R(t) - F(t) \alpha_c (T(t)-T_c), \label{eq:PINN_model_Tdot}
    	\end{align}
    \end{subequations}    
    with the \textit{unknown} and unmeasured reaction rate $R(t) = c(t)ke^{-\frac{N}{T(t)}}$.

    The goal is to minimize production costs. To enable flexible operation taking advantage of electricity price fluctuations, we assume the existence of a product storage with filling level $l$ and a maximum capacity of six hours of steady-state production. Given electricity price predictions, the controller aims to minimize production costs while ensuring that a steady product demand is met and adhering to bounds imposed on the system states. Production costs can be influenced by altering the process cooling as the electric power consumption is assumed to be proportional to the coolant flow rate $F$. Table \ref{tab:CSTR_bounds_and_ss} presents the state bounds and the steady-state values of the model (see Eq. \eqref{eq:CSTR_model}). 
    \begin{table}[ht]
    \centering
    \setlength{\extrarowheight}{0.05cm}
        \caption{Lower (lb) and upper (ub) bounds of system states and control inputs and steady-state (ss) values used to evaluate the economic benefit of flexible production in eNMPC.}
        \label{tab:CSTR_bounds_and_ss}
        \begin{tabular}{l|rrr}
            \toprule
            variable              & lb                & ub                  & ss        \\ \hline
            $c$                   & 0.1231            & 0.1504              & 0.1367    \\
            $T$                   & 0.6               & 0.8                 & 0.7293    \\
            $\rho$                & $0.8\frac{1}{h}$  & $1.2\frac{1}{h}$    & $1.0\frac{1}{h}$ \\
            $F$                   & $0.0\frac{1}{h}$  & $700.0\frac{1}{h}$  & $390.0\frac{1}{h}$ \\
            \bottomrule
        \end{tabular}
    \end{table}
    Matching the hourly structure of the day-ahead electricity market, we choose control steps of length $\Delta t_{\text{ctrl}} = \SI{1}{\text{h}}$. We use historic day-ahead electricity prices from the Austrian market (\cite{open_power_system_data_2020}). During training, we use the prices from March 29, 2015 to March 25, 2018. For the final evaluation of the trained policies, we use the prices from March 26, 2018 to September 30, 2018.

\subsection{Implementation details}\label{sec:results_training}
    This subsection explains the implementation details of our method (Sec. \ref{sec:method_PIMBRL-Koopman}) when it is applied to this case study. Sec. \ref{sec:results_training_architecture} presents our architecture choices for the agent and the dynamic models. Thereafter, three subsections address the iterative three-step approach of our method, i.e., data sampling (Sec. \ref{sec:results_training_datasampling}), system identification (Sec. \ref{sec:results_training_SI}), and policy optimization (Sec. \ref{sec:results_training_RL}). Sec. \ref{sec:results_training_benchmarks} briefly describes alternative methods for learning a controller, e.g., learning a neural network policy via (physics-informed) MBPO, which we use to rate the performance of our proposed method. All training code including the hyperparameters that were used to obtain the results presented in Section \ref{sec:results_results} is available online\footnote{\url{https://jugit.fz-juelich.de/iek-10/public/optimization/pi-mbpo4koopmanenmpc}}.

\subsubsection{Model architecture}\label{sec:results_training_architecture}
    As in \cite{mayfrank2024KoopmanPPO}, we choose a latent space dimensionality of eight for the Koopman model that is part of the eNMPC policy. Since the CSTR model has two states and two control inputs (see Eq. \eqref{eq:CSTR_model}), this means that $\bm{A}_{\bm{\theta}_\text{K}} \in \mathbb{R}^{8 \times 8}, \bm{B}_{\bm{\theta}_\text{K}} \in \mathbb{R}^{8 \times 2}, \bm{C}_{\bm{\theta}_\text{K}} \in \mathbb{R}^{2 \times 8}$ (see Eq. \eqref{eq:Koopman_model}). The encoder $\bm{\psi}_{\bm{\theta}_\text{K}}\colon \mathbb{R}^2 \mapsto \mathbb{R}^8$ is a multilayer perceptron (MLP) with two hidden layers (four and six neurons, respectively) and hyperbolic tangent activation functions.

    RL-based training of (e)NMPC controllers requires solving and differentiating through many OCP instances. Since the number of variables in an OCP grows linearly with the number of time steps, RL training of (e)NMPCs is computationally challenging given long prediction horizons. To balance control performance and computational tractability, we determine an effective eNMPC prediction horizon by repeatedly solving eNMPC problems using the mechanistic CSTR model while varying the prediction horizon. We select a horizon $t_f$ of nine hours, as longer horizons do not produce substantial performance gains in the mechanistic eNMPC. Thus, $\mathrm{T}_{+1} = \{ t,\dots,t+9 \}$ and $\mathrm{T} = \{ t,\dots,t+8 \}$ (see Eq. \eqref{eq:Koopman_OCP_v2}). Analogous to our earlier work on model-free RL of Koopman eNMPCs (\cite{mayfrank2024KoopmanPPO}), given a prediction for the evolution of the electricity prices $\bm{p}_{\text{eNMPC}} = (p_t)_{t \in \mathrm{T}_{+1}}$, the policy aims to minimize the production cost while satisfying the bounds of the states and the product storage. To that end it first calculates the initial latent state $\bm{z}_0$ by passing the initial state $\bm{x}_0 = (c_0, T_0)^\top$ through the encoder, i.e., $\bm{z}_0 = \bm{\psi}_{\bm{\theta}_\text{K}}(\bm{x}_0)$. Then, the following OCP is solved:
    \begin{subequations}\label{eq:Koopman_eNMPC}
    \allowdisplaybreaks
	\begin{align}
		\underset{(\rho_t, F_t)_{t \in \mathrm{T}}}{\min} &\sum_{t \in \mathrm{T}_{+1}} (F_t p_t \Delta t_{\text{ctrl}} + M\bm{s}^{\intercal}_{t}\bm{s}_t),\\
		\text{s.t. }\bm{z}_{t+1} &= \bm{A}_{\bm{\theta}_\text{K}}\bm{z}_t + \bm{B}_{\bm{\theta}_\text{K}}\bm{u}_t \quad\forall t \in \mathrm{T},\\
		l_{t+1} &= l_t + (\rho_t - \rho_{\text{ss}}) \Delta t_{\text{ctrl}} \quad\forall t \in \mathrm{T},\\
		\bm{x}_{t} &= \bm{C}_{\bm{\theta}_\text{K}}\bm{z}_t \quad\forall t \in \mathrm{T}_{+1},\label{eq:Koopman_eNMPC_decode}\\
		\underline{\bm{x}}_t - \bm{s}_{\bm{x},t} + \bm{\theta}_{\text{B},\underline{\bm{x}}} &\leq \bm{x}_t \leq \bar{\bm{x}}_t + \bm{s}_{\bm{x},t} + \bm{\theta}_{\text{B},\bar{\bm{x}}} \quad\forall  t \in \mathrm{T}_{+1},\label{eq:Koopman_eNMPC_StateBounds}\\
		0 - s_{l,t} + \theta_{\text{B},\underline{l}} &\leq l_t \leq 6.0 + s_{l,t} + \theta_{\text{B},\bar{l}} \quad\forall  t \in \mathrm{T}_{+1},\label{eq:Koopman_eNMPC_StorageBounds}\\
		\bm{s}_t &= \colvec{2}{\bm{s}_{\bm{x},t}}{s_{l,t}} \quad\forall  t \in \mathrm{T}_{+1},\\
		\bm{0} &\leq \bm{s}_t \quad\forall  t \in \mathrm{T}_{+1},\\
		\underline{\bm{u}}_t &\leq \bm{u}_t \leq \bar{\bm{u}}_t \quad\forall  t \in \mathrm{T}
	\end{align}
    \end{subequations}
    Note that $\bm{\theta}_{\text{B}}$ appears only in the constraints regarding the state bounds of the CSTR (Eq. \eqref{eq:Koopman_eNMPC_StateBounds}) and the storage level (Eq. \eqref{eq:Koopman_eNMPC_StorageBounds}), i.e., $\bm{\theta}_{\text{B}}$ merely serves to tighten or relax those bounds.

\begin{figure}[ht]
    	\centering
    	\includegraphics[width=0.4\paperwidth]{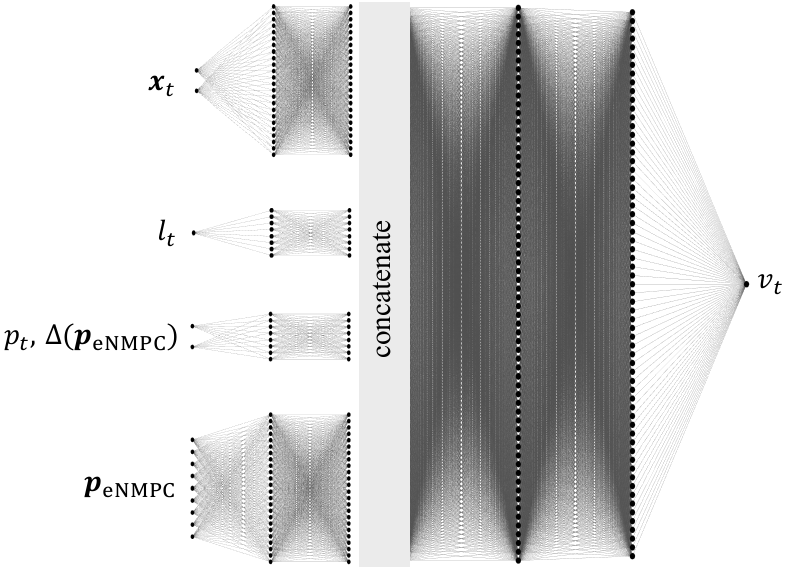}
    	\caption{Critic architecture. As for the policy, the system states are scaled so that the feasible range is in [-1,1], whereas the product storage and the electricity prices are left unscaled.}
    	\label{fig:critic_nn}
    \end{figure}

    We use a NN with parameters $\bm{\phi}$ as the critic in the policy optimization step. The architecture of the critic is shown in Fig. \ref{fig:critic_nn}. It has four separate input layers for (i) $\bm{x}_t$, (ii) $l_t$, (iii) a two-element vector that includes the electricity price at the current time step and the difference between the highest and the lowest electricity price in the current MPC prediction horizon, i.e., $\Delta(\bm{p}_{\text{eNMPC}}) = \max_{t \in \mathrm{T}_{+1}} (p_t) - \min_{t \in \mathrm{T}_{+1}} (p_t)$, and (iv) a vector of all electricity prices in the current MPC prediction horizon. Each of those input layers is followed by two equally sized hidden layers, with 24, 8, 8, and 24 neurons, respectively. The output of all second hidden layers is then concatenated and passed through two fully connected layers, each of size 64 neurons. The output layer has a single neuron for the value of the current state. Except the output layer, which does not have an activation function, all layers have hyperbolic tangent activation functions. We pick such an architecture since it yielded substantially better results than fully connected architectures of similar overall size in preliminary testing.

\begin{figure*}[ht]
    \centering
    \includegraphics[width=0.7\paperwidth]{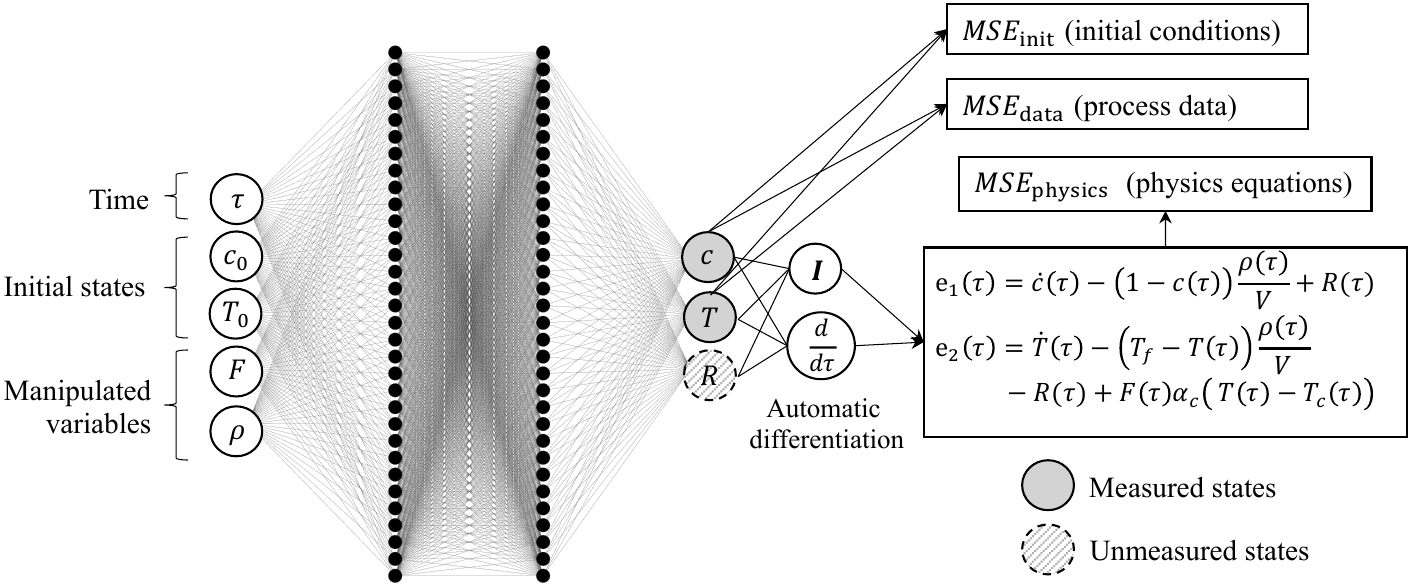}
    \caption{General schematic of the PINN models used in the CSTR case study.}
    \label{fig:pinn-model}
    \end{figure*}

    In the policy optimization step, we use an ensemble of $n=10$ PINNs to model the dynamics of the real environment. Each PINN $\bm{NN}_{i,\bm{\omega}_i} \forall i \in \{1,2,\dots,10\}$ has five separate input features that are concatenated to a single input layer (i) the PINN time $\tau$, (ii) two initial states $c_0$, $T_0$, and (iii) two control inputs $\rho$, $F$. Here, the PINN time is chosen to be $\tau \in [0, \Delta t_{\text{ctrl}}]$ such that the PINN can be trained with constant control inputs and the PINN time domain matches the size of a control step. The input layer is followed by two equal-sized hidden layers with 32 neurons each. The output layer consists of three output features: two for the differential states $\bm{x} = [c, T]^\intercal$ and one for the algebraic state $\bm{y} = [R]$. A schematic of the PINN architecture is shown in Fig. \ref{fig:pinn-model}. Further details on the PINN modeling approach used throughout this paper can be found in \cite{VELIOGLU2025108899}.

    At the beginning of a training run, we initialize $\bm{\theta}_{\text{B}}$ with zeros, i.e., our initial guess is that the bounds do not need to be adapted. The learnable parameters of the PINN models ($\bm{\omega}$) are initialized with the Xavier normal distribution (\cite{pmlr-v9-glorot10a}) since we observed that it leads to better performance for PINNs in our preliminary studies. All other learnable parameters ($\bm{\theta}_{\text{K}}, \bm{\phi}$) are initialized randomly using the default PyTorch (\cite{paszke2017automatic}) parameter initialization method.

    For the purpose of data-driven modeling, we rescale the system states and control inputs (see Table \ref{tab:CSTR_bounds_and_ss}) linearly so that the lower and upper bounds of each variable correspond to $-1.0$ and $1.0$, respectively.

\subsubsection{Data sampling}\label{sec:results_training_datasampling}
    Each iteration of the MBPO algorithm (\cite{Janner2019MBPO}) starts with the current policy interacting with the environment to gather data about the system dynamics (first step in Fig. \ref{fig:method_mbpo}). This data is then added to the data set $\mathcal{D}$. We gradually increase the number of steps that the policy takes at each MBPO iteration: In the first 10 iterations, we let the policy take 20 steps in each iteration, i.e., until $\mathcal{D}$ contains 200 steps. Then, until up to 500 overall steps, we increase the number of steps taken at each iteration to 50. Finally, we increase this number to 250 steps per iteration until we reach an overall number of 2500 steps in $\mathcal{D}$. Here, we terminate each training run.

    In each MBPO iteration, we start the data sampling step by resetting the environment, i.e., we set the system states to their steady-state values, we randomly initialize the storage filling level between one and two hours of steady-state production, and we sample a series of electricity prices for the current episode. An episode ends given one of two conditions: (i) the episode reaches its maximum number of 167 steps, i.e., one week of uninterrupted closed-loop operation, or (ii) a constraint violation occurs in one of the states, where the distance of the variable from the violated bound is bigger than the feasible interval of the associated state (see Tab. \ref{tab:CSTR_bounds_and_ss}). Upon termination of an episode, the environment resets and a new episode starts. All state transitions are added to $\mathcal{D}$. Data sampling continues until the desired number of steps in the environment in the current MBPO iteration has been reached.

    To enable early stopping in the SI step (see Sec. \ref{sec:results_training_SI}), we split $\mathcal{D}$ into a training and a validation data set, i.e., $\mathcal{D} = (\mathcal{D}_\text{train}, \mathcal{D}_\text{val})$. In the first MBPO iteration, the data from the first episode is added to $\mathcal{D}_\text{train}$, and the data from the second episode to $\mathcal{D}_\text{val}$. Thereafter, at the start of each episode, we randomly determine whether the data of this episode will be added to $\mathcal{D}_\text{train}$ (with a chance of \SI{75}{\%}) or $\mathcal{D}_\text{val}$ (\SI{25}{\%} chance).

    In the data sampling step of the first MBPO iteration, the policy still has randomly initialized parameters $\bm{\theta}_{\text{K}}$. Herein, we, therefore, randomly sample the actions $\bm{u}_t$ from a uniform distribution over the action space. In all subsequent MBPO iterations, we sample the action $\bm{u}_t$ from a normal distribution $\bm{u}_t \sim \bm{\mathcal{N}}(\bm{u}^*_t,\bm{\sigma}^{2})$, where $\bm{u}^*_t$ is the deterministic output of the policy. At the start of each new episode, $\bm{\sigma}$ is randomly sampled from a uniform distribution between $0.0$ and $0.1$.

    The PINN requires two additional data sets for training: (i) A physics data set $\mathcal{D}_\text{physics}$ is used to calculate the physics residuals. It contains unlabeled data, i.e., no output observations from the environment are needed. We sample $\lvert \mathcal{D}_\text{physics} \rvert = 2000 $ collocation points for the PINN inputs using the lower and upper bounds specified in Table \ref{tab:CSTR_bounds_and_ss} for the initial states and controls, and $\tau \in [0, \Delta t_{\text{ctrl}} = \SI{1}{\hour}]$ for the PINN time. (ii) An initial state data set $\mathcal{D}_\text{init}$ is used to teach the PINN to match the initial state at $t = t_0$. Although this data set is labeled, the output states are identical to the initial states. Thus, no interaction with the environment is required to assemble $\mathcal{D}_\text{init}$. We sample $ \lvert \mathcal{D}_\text{init} \rvert = 100$ data points for the state and control variables within the bounds specified in Table \ref{tab:CSTR_bounds_and_ss}. Both $\mathcal{D}_\text{physics}$ and $\mathcal{D}_\text{init}$ are generated uniquely for each PINN model $\bm{NN}_{i,\bm{\omega}_i} \forall i \in \{1,2,\dots,10\}$ in the ensemble but remain unchanged throughout the training. We use Latin Hypercube Sampling (LHS) (\cite{Iman1981AnAssessment}) to ensure coverage of the PINN input domain.
    
\subsubsection{System identification}\label{sec:results_training_SI}
    In the SI step (second step in Fig. \ref{fig:method_mbpo}), we fit the trainable parameters of the Koopman model ($\bm{\theta}_{\text{K}}$) and of the model ensemble ($\bm{\omega}_i \forall i \in \{1,2,\dots,10\}$) to the data in $\mathcal{D}_\text{train}$.

    For the Koopman model, we use the Adam optimizer (\cite{kingma2014adam}) with a learning rate of $10^{-4}$ and a mini-batch size of 64 samples. We choose a maximum number of 5000 epochs; however, we stop SI early if the sum of Eqs. \eqref{eq:koopman_SI_losses_AE} - \eqref{eq:koopman_SI_losses_comb} with respect to $\mathcal{D}_\text{val}$ does not reach a new minimum for 25 consecutive epochs.

    The PINN models in the ensemble are trained by minimizing the following loss function, where $n$ is the number of states:
    \begin{subequations}
    \label{eq:pinn-loss-ic-cv}
    \begin{align}
    MSE_{\text{total}} &=  MSE_{\text{physics}} + \lambda_1 MSE_{\text{data}} + \lambda_2 MSE_{\text{init}}, \label{eq:totalloss} \\
    \label{eq:data-loss}\text{with } MSE_{\text{data}} &= \frac{1}{n \lvert \mathcal{D}_\text{train} \rvert} \sum_{j=1}^{\lvert \mathcal{D}_\text{train} \rvert} (\Vxt(\tau_j) -  \Vx(\tau_j))^2, \\
    \begin{split}
    \label{eq:res-loss}MSE_{\text{physics}}  &= \frac{1}{n \lvert \mathcal{D}_\text{physics} \rvert} \sum_{j=1}^{\lvert \mathcal{D}_\text{physics} \rvert} \left(\Dot{\Vxt}(\tau_j) - {\bm f}(\Vxt(\tau_j), \Vyt(\tau_j), \Vu_j)\right)^2,         
    \end{split} \\
    \label{eq:init-loss}MSE_{\text{init}} &= \frac{1}{n \lvert \mathcal{D}_\text{init} \rvert} \sum_{j=1}^{\lvert \mathcal{D}_\text{init} \rvert} (\Vxt_j(0) - \Vx_j(0))^2
    \end{align}
    \end{subequations}
    Here, $MSE_{\text{data}}$ corresponds to the loss term for measurement data, $MSE_{\text{physics}}$ corresponds to the physics regularization loss stemming from (incomplete) physics knowledge on system dynamics (c.f. Eq. \eqref{eq:PINN_model}), and $MSE_{\text{init}}$ corresponds to a loss term that ensures that the predictions at $\tau = 0$ are consistent with the initial states. $\lambda_1$ and $\lambda_2$ denote the weights of the measurement data and initial condition loss terms. 

    The PINN models are trained in a two-stage manner, similar to \cite{VELIOGLU2025108899}. In the first stage, we use the Adam optimizer (\cite{kingma2014adam}) for 1000 epochs, with a learning rate of $10^{-3}$ and a mini-batch size of 64 samples. Here, we use inverse Dirichlet weighting (\cite{Maddu2022InverseNetworks}) to obtain the weights $\lambda_1$, $\lambda_2$ in Eq. \eqref{eq:totalloss} dynamically. In the second stage, we use the LBFG-S optimizer (\cite{Liu1989OnOptimization}) with a full batch for a maximum number of 300 epochs; however, we stop early if the loss in Eq. \eqref{eq:pinn-loss-ic-cv} with respect to $\mathcal{D}_\text{val}$ does not reach a new minimum for 25 consecutive epochs. Note that in the second stage, we fix the weights $\lambda_1$, $\lambda_2$ according to the last value attained in the first stage. At each MBPO iteration, the trainable parameters of the model ensemble ($\bm{\omega}_i \, \forall \, i \in \{1,2,\dots,10\}$) have a $1/3$ chance of resetting to prevent getting stuck in a sub-optimal local minimum over many consecutive iterations.

\subsubsection{Policy optimization}\label{sec:results_training_RL}
    The policy optimization step (third step in Fig. \ref{fig:method_mbpo}) adjusts the parameters $\bm{\theta}_{\text{B}}$ towards task-optimal performance of the eNMPC policy, given the Koopman model that was identified through SI. Using the PINN ensemble, we construct a data-driven surrogate RL environment. In each step taken in this surrogate environment, one of the PINNs is chosen randomly as an approximation of the state transition function (Eq. \eqref{eq:transition_function}) by evaluating it at $\tau = \Delta t_{\text{ctrl}}$. As is typical practice in MBPO (\cite{Janner2019MBPO}), we shorten episodes in the surrogate environment (to a maximum of eight steps) to prevent compounding prediction errors of the PINNs from negatively influencing the policy optimization. As in the data sampling step (see Sec. \ref{sec:results_training_datasampling}), we still terminate an episode earlier whenever an outsized constraint violation occurs. Whenever an episode ends and the surrogate environment resets, we randomly sample the state of the CSTR from all states in $\mathcal{D}_\text{train}$, thus decoupling the length of simulated episodes from the state values that can be reached during an episode.

    To incentivize the desired controller behavior, we choose a reward that promotes cost savings compared to steady-state production while punishing constraint violations. The overall reward at each step (Eq. \eqref{eq:reward_function}) is calculated via
    \begin{equation}\label{eq:our_reward_function}
        r_{t} = \alpha \cdot r^{\text{cost}}_{t} - r^{\text{con,rel}}_{t} - r^{\text{con,bool}}_{t} + 1.
    \end{equation}
    Herein, $r^{\text{cost}}_{t}$ incentivizes the minimization of the production costs by giving positive rewards if the costs are lower than those of a steady-state production regime:
    \begin{equation*}
        r^{\text{cost}}_{t} = (F_{\text{ss}} - F_{t-1}) \cdot p_{t-1} \cdot \Delta t_{\text{ctrl}}
    \end{equation*}
    Constraint violations are penalized twofold: $r^{\text{con,rel}}_{t}$ penalizes constraint violations quadratically, i.e., $r^{\text{con,rel}}_{t} \geq 0$, and $r^{\text{con,rel}}_{t} = 0$ if no constraint violation occurs at $t$. $r^{\text{con,bool}}_{t}$ imposes an additional small constant penalty whenever a constraint violation occurs, irrespective of the magnitude of the violation, i.e., $r^{\text{con,bool}}_{t} = 0.1$ if there is a constraint violation, and $r^{\text{con,bool}}_{t} = 0$ otherwise. At every step, we add a constant reward of $1$ to ensure that the sum of rewards of an episode keeps rising as long as the episode continues, i.e., we give a reward for not producing a constraint violation that is large enough to cause an environment reset. $\alpha = 5 \cdot 10^{-6}$ is a hyperparameter used to balance the influence of $r^{\text{cost}}_{t}$ compared to all other components of the overall reward.

    We use our previously published method for automatic differentiation of Koopman (e)NMPCs (\cite{mayfrank2024KoopmanPPO}) in conjunction with the \textit{Stable-Baselines3} (\cite{stable-baselines3}) implementation of the PPO algorithm (\cite{schulman2017proximal}) for policy optimization. In each PPO iteration, we sample $2048$ steps in the surrogate environment, and we set the batch size for the policy and critic updates to $256$ samples. We use the Adam optimizer (\cite{kingma2014adam}) with a learning rate of $10^{-3}$, and we clip the gradient norms of the policy and the critic to a maximum of $0.5$.

    As explained in Sec. \ref{sec:method_PIMBRL-Koopman}, we do not terminate the policy optimization step after a predefined number of PPO iterations. Instead, following the idea of \cite{kurutach2018model}, we implement the following performance-based stopping criterion: After every five iterations of the PPO algorithm, we evaluate the performance of the policy separately on all 10 learned models. To this end, we set up 10 validation environments corresponding to the 10 learned models. In each of these environments, only the corresponding model is used to represent the transition function (Eq. \eqref{eq:transition_function}). Then, we compute the ratio of validation environments in which the policy improves by running the policy for five episodes in each environment. Specifically, we check the ratio of validation environments in which the policy has reached a new highest average reward in the last 25 PPO iterations. When this ratio falls below $70\%$, we terminate policy optimization. Then, the next MBPO iteration begins with the data sampling step, or the overall training process stops if we have already reached 2500 steps in the real environment.

\subsubsection{Ablation and benchmark variants}\label{sec:results_training_benchmarks}
    \newcommand{\SIKoopPIRLBounds}{$\text{SI}_{\text{Koop}}\text{PIRL}_{\text{Bounds}}$}
    \newcommand{\SIKoopRLBounds}{$\text{SI}_{\text{Koop}}\text{RL}_{\text{Bounds}}$}
    \newcommand{\SIKoop}{$\text{SI}_{\text{Koop}}$}
    \newcommand{\PIRLMLP}{$\text{PIRL}_{\text{MLP}}$}
    \newcommand{\RLMLP}{$\text{RL}_{\text{MLP}}$}
    To evaluate the performance of our proposed method and to analyze how different components of the method contribute to the overall performance, we compare the performance of the following controller types. To avoid unnecessary repetition, we focus our description on the differences compared to the main method, e.g., unless explicitly stated otherwise, all variants use MBPO (\cite{Janner2019MBPO}).
    \begin{enumerate}
        \item \SIKoopPIRLBounds{} (main contribution): Our method, outlined in Sec. \ref{sec:method_PIMBRL-Koopman} with implementation details explained in Sec. \ref{sec:results_training_architecture} - \ref{sec:results_training_RL}. The name refers to the iterative process of SI of a Koopman model, followed by physics-informed (PI) RL of the bounds in the eNMPC.
        \item \SIKoopRLBounds{}: Here, we train the model ensemble without assuming any prior physics knowledge. Each model in this ensemble is a vanilla NN. We train the models by minimizing the discrete-time $\text{L}_2$ prediction loss, i.e., given $\mathcal{D}$, we minimize $|| \bm{NN}_{i,\bm{\omega}_i} (\bm{x}_t,\bm{u}_t) - \bm{x}_{t+1} ||$ for each $i \in \{1,2,\dots,10\}$. The architecture of the models is identical to that of the PINN models (see Fig. \ref{fig:pinn-model}), except that we remove the time $\tau$ from the inputs and the reaction rate $R$ from the outputs.
        \item \SIKoop{}: This variant can be thought of as adaptive Koopman eNMPC. Compared to \SIKoopPIRLBounds{}, we keep the data sampling step and the iterative SI of the Koopman model unchanged, but we omit any further (model-based) policy optimization.
        \item \PIRLMLP{}: Here, we use a neural network policy in form of an MLP instead of a Koopman eNMPC. This policy has the same architecture as the critic described in Sec. \ref{sec:results_training_architecture} (Fig. \ref{fig:critic_nn}), except that its output layer has a size of two, corresponding to the two-dimensional action space of the environment.
        \item \RLMLP{}: Same as \PIRLMLP{} but without physics-informed model training.
    \end{enumerate}

\subsection{Results}\label{sec:results_results}
    For each type of controller, we repeat the training ten times using different fixed seeds in every training run. We train each controller type for 2500 steps in the real environment as specified in Sec. \ref{sec:results_training}. After every full MBPO iteration, we save the agent and the model ensemble for testing purposes.

    \begin{figure*}[!htb]
        \centering
        \subfloat{\includegraphics[width=0.38\paperwidth]{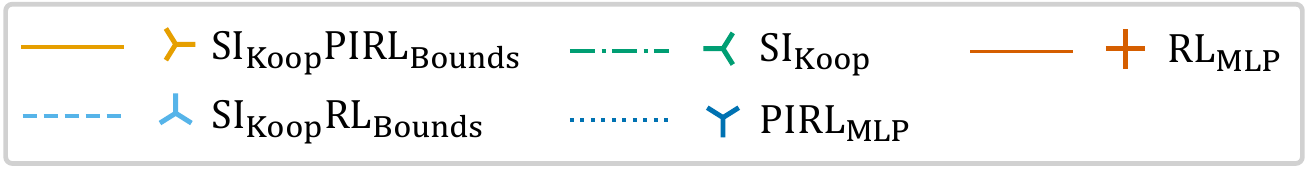} \label{fig:results_plot_legend}} \\
        \setcounter{subfigure}{0} 
        \subfloat[a][Average sum of rewards obtained in each test episode.]{\includegraphics[width=0.36\paperwidth]{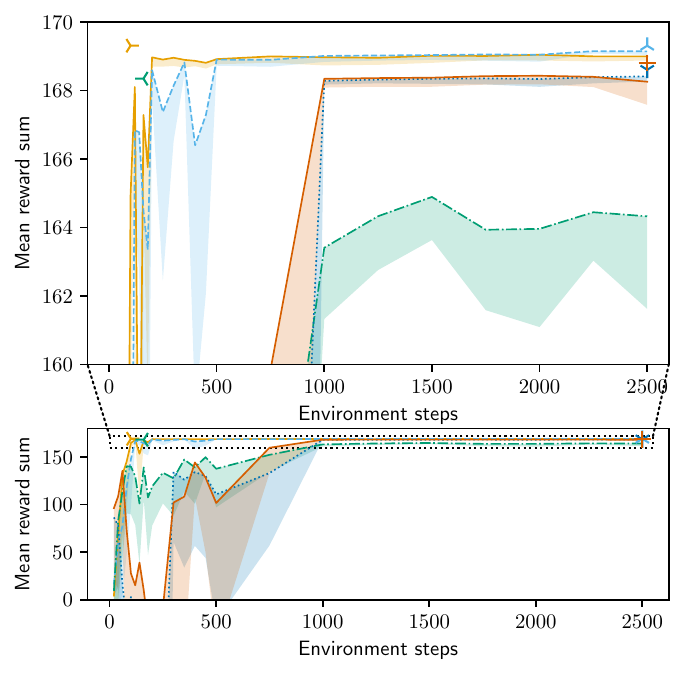} \label{fig:trainingresults_rewards}} \hspace{10pt}
        \setcounter{subfigure}{2} 
        \subfloat[c][Average number of constraint violations produced in each test episode.]{\includegraphics[width=0.36\paperwidth]{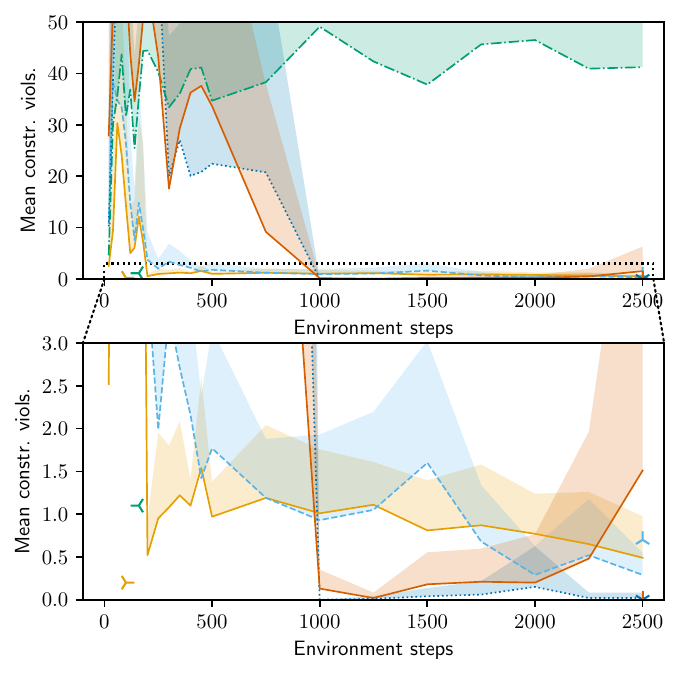} \label{fig:trainingresults_constrviols}} \\
        \setcounter{subfigure}{1} 
        \subfloat[b][Variance of the sum of rewards across differently seeded training runs.]{\includegraphics[width=0.36\paperwidth]{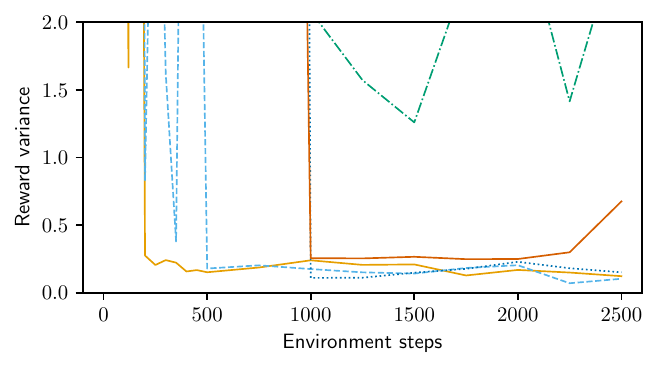} \label{fig:trainingresults_rewardvariance}} \hspace{10pt}
        \setcounter{subfigure}{3} 
        \subfloat[d][Average economic cost incurred in each test episode.]{\includegraphics[width=0.36\paperwidth]{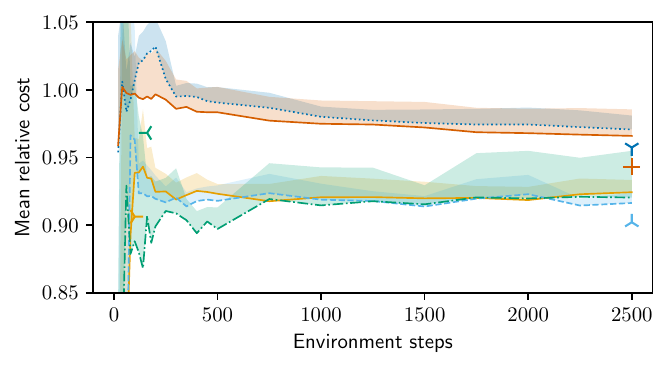} \label{fig:trainingresults_costs}}
        \caption{Performance metrics for different controller types. Each line represents the mean metric across 10 test episodes, averaged over 10 controllers trained with different random seeds. Shaded areas denote one standard deviation from the mean, shown in a single direction (toward worse performance) for clarity. We add a point marker for each controller type to indicate the value obtained by the controller that achieved the highest average reward over the 10 test episodes.}
        \label{fig:trainingresults}
    \end{figure*}
\begin{figure*}[!htb]
        \centering
        \includegraphics[width=0.27\paperwidth]{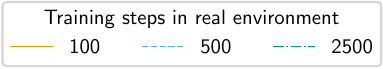} \label{fig:trajectories_legend}\vspace{3pt} \\
        \SIKoopPIRLBounds{}\vspace{2pt} \\ 
        \includegraphics[width=0.75\paperwidth]{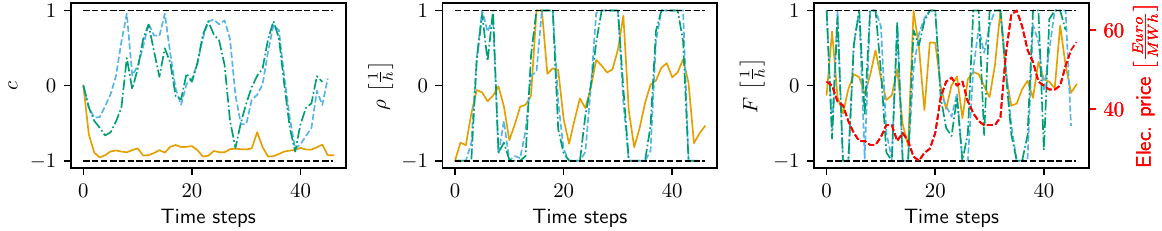} \label{fig:trajectories_Bounds_PIRL}\vspace{3pt} \\
        \SIKoopRLBounds{}\vspace{2pt} \\ 
        \includegraphics[width=0.75\paperwidth]{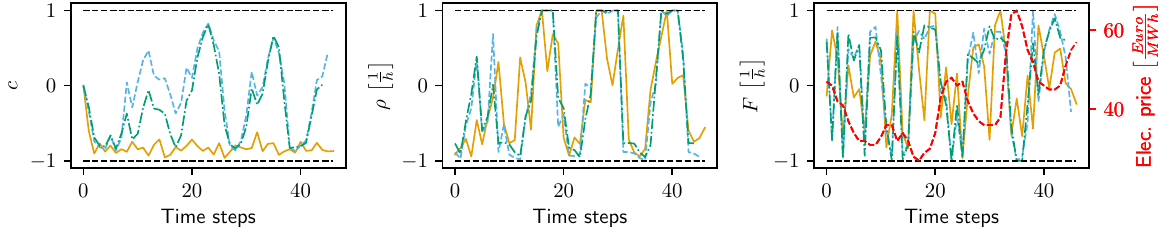} \label{fig:trajectories_Koopman_PIRL}\vspace{3pt} \\
        \SIKoop{}\vspace{2pt} \\ 
        \includegraphics[width=0.75\paperwidth]{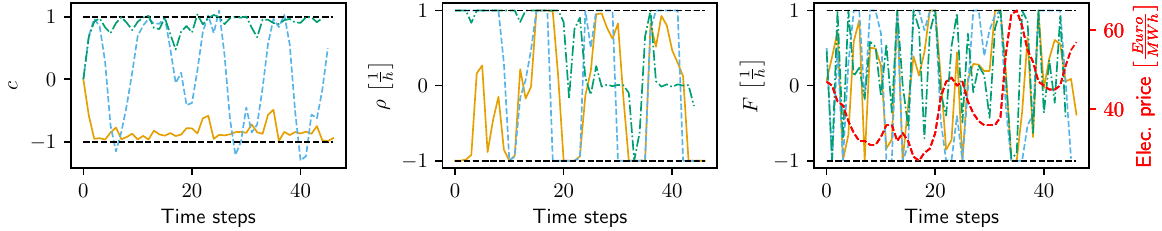} \label{fig:trajectories_Koopman_SI}\vspace{3pt} \\
        \PIRLMLP{}\vspace{2pt} \\ 
        \includegraphics[width=0.75\paperwidth]{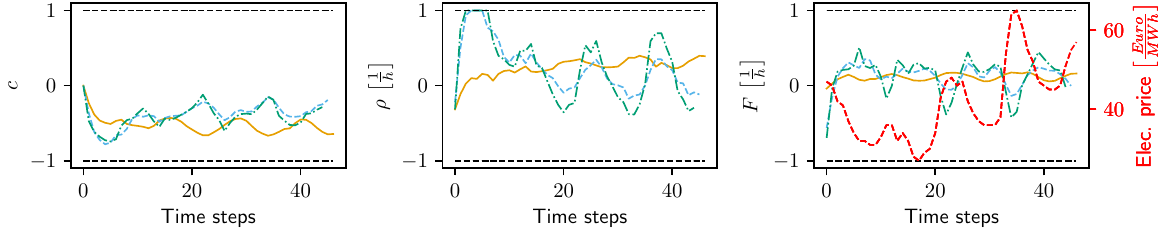} \label{fig:trajectories_PI_MLP} \\
        \RLMLP{}\vspace{2pt} \\ 
        \includegraphics[width=0.75\paperwidth]{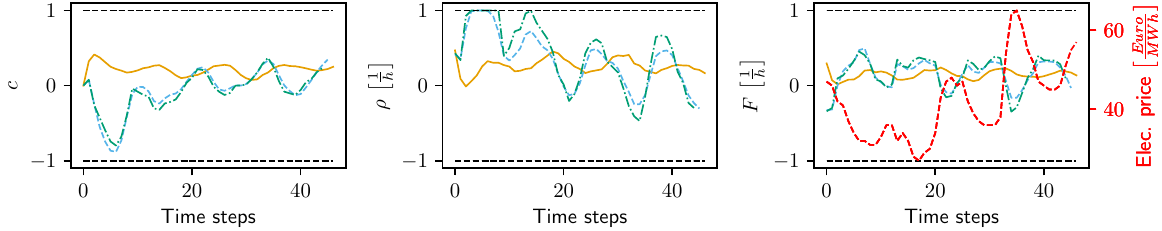} \label{fig:trajectories_MLP} \\
        \caption{Control trajectory comparison. The bounds of each variable (see Tab. \ref{tab:CSTR_bounds_and_ss}) are used for scaling to the [-1, 1] range. We omit the temperature $T$ since it never reaches its bounds in any of the test episodes.}
        \label{fig:control_trajectories}
    \end{figure*}

    We test each controller by running it without exploration noise, i.e., $\bm{u}_t = \bm{u}^*_t$, for 10 one-week-long episodes (168 steps in each episode) using electricity price trajectories that were not used during training. Each test is performed using the same 10 electricity price trajectories, ensuring comparable results between the tests. The aggregated results of these tests can be viewed in Fig. \ref{fig:trainingresults}. Fig. \ref{fig:control_trajectories} shows some randomly chosen control trajectories which were part of these tests. Since our focus is on sample efficiency rather than final policy performance, we show control performance at different numbers of environment steps. We randomly select one training run for each of the depicted controller types and show its control performance in Fig. \ref{fig:control_trajectories}. Therein, we randomly select one of the 10 test electricity price trajectories and show the first 48 time steps of that test episode for each controller type.

    To evaluate the performance of the controllers, we analyze the obtained rewards (since this is the metric that is maximized by MBPO), and the two metrics which we are primarily interested in and which together produce the reward (see Eq. \eqref{eq:our_reward_function}), i.e., the constraint violations and the economic performance. Throughout this section, we report economic performance via the economic cost incurred relative to the nominal production cost, i.e., we report the total cost incurred by the respective controller divided by the cost of steady-state production at nominal rate given the same electricity price trajectory. Fig. \ref{fig:trainingresults_rewards} shows that \SIKoopPIRLBounds{} and \SIKoopRLBounds{} (i) reach the highest reward values and that they do so with (ii) exceptional sample efficiency and (iii) low variance (see also Fig. \ref{fig:trainingresults_rewardvariance}) across different training runs. \PIRLMLP{} and \RLMLP{} also achieve low performance variance across training runs, albeit with lower sample efficiency and at a lower performance level (when measured in average rewards) compared to \SIKoopPIRLBounds{} and \SIKoopRLBounds{}. \SIKoop{} performs worse, not converging to high and stable rewards within the given budget of 2500 environment steps.

    A direct comparison of the physics-informed vs.~the purely data-driven variants shows a small but consistent benefit resulting from the utilization of partial physics knowledge in training the model ensemble: \SIKoopPIRLBounds{} performs better than \SIKoopRLBounds{} between 200 and 500 environment steps. Thereafter, both variants perform comparably well. The two MLP controller variants,  \PIRLMLP{} and \RLMLP{}, show similar sample efficiency. Still, the PINN ensemble seems to benefit the stability of the learning once relatively high rewards have been reached.
    
    Looking at Fig. \ref{fig:trainingresults_constrviols}, we can see that initially (up to around 1000 steps) \SIKoopPIRLBounds{} and \SIKoopRLBounds{} are best at avoiding constraint violations. In particular, \SIKoopPIRLBounds{} quickly learns to avoid constraint violations. Both variants continue to make relatively steady progress in this regard up until the end of the training runs. The MLP controllers initially cause many constraint violations, however, after around 1000 environment steps they have learned to avoid constraint violations almost perfectly. Towards the end of the training, some of the \RLMLP{} controllers seem to loose this capability partially. \SIKoop{} struggles with constraint satisfaction across the full length of all training runs.

    Fig. \ref{fig:trainingresults_costs} shows that, when looking at the economic performance, there is a clear difference between the Koopman eNMPC-based controllers and the MLP controllers: After around 750 environment steps, the Koopman eNMPC-based controllers all produce average costs of around 91 \% to 93 \% of nominal production costs. After that, no big improvement happens. The MLP controllers incur substantially higher production costs (between 96 \% and 102 \% after 750 environment steps); however, they keep improving until the end of the training. Thus, it is possible that their economic performance would eventually become comparable to that of the Koopman eNMPC-based controllers if given a higher training budget.

    The control trajectories in Fig. \ref{fig:control_trajectories} align with the findings derived thus far from Fig. \ref{fig:trainingresults}. All controllers show an intuitive inverse relationship between electricity prices and coolant flow rate $F$, although this relationship is noticeably weaker for \PIRLMLP{} and \RLMLP{}. Moreover, the MLP controllers do not utilize the full range of the control inputs frequently. These observations match the lower cost savings of the MLP controllers. Regarding the evolution of the product concentration $c$, \SIKoopPIRLBounds{} and \SIKoopRLBounds{} effectively utilize the full feasible range without violating bounds. In contrast, \SIKoop{} causes minor constraint violations. \PIRLMLP{} and \RLMLP{} avoid violations by maintaining $c$ well within bounds but sacrifice process flexibility, limiting economic performance.

    \begin{figure*}[!htb]
        \centering
        \subfloat{\includegraphics[width=0.3\paperwidth]{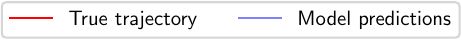} \label{fig:prediction_test_legend}} \\
        \setcounter{subfigure}{0} 
        \subfloat[a][]{\includegraphics[width=0.25\paperwidth]{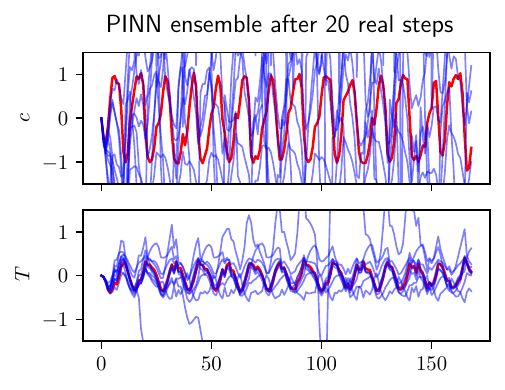} \label{fig:predictions_PINN_20}}
        \subfloat[b][]{\includegraphics[width=0.25\paperwidth]{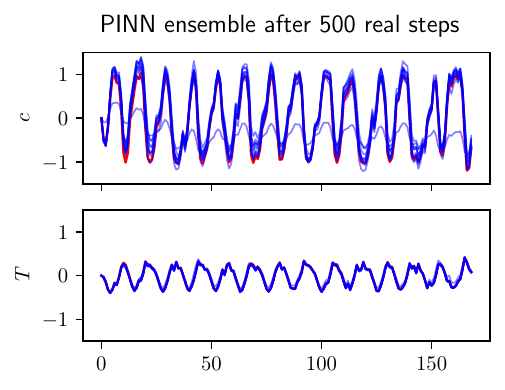} \label{fig:predictions_PINN_500}}
        \subfloat[c][]{\includegraphics[width=0.25\paperwidth]{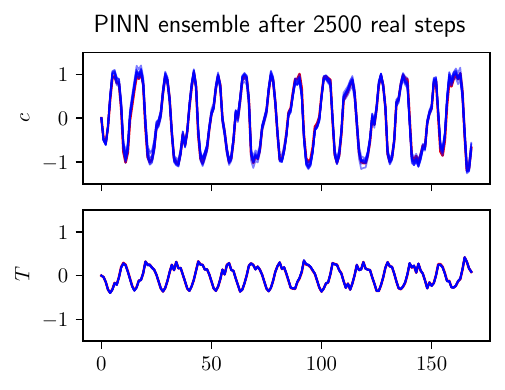} \label{fig:predictions_PINN_2500}} \\
        \subfloat[d][]{\includegraphics[width=0.25\paperwidth]{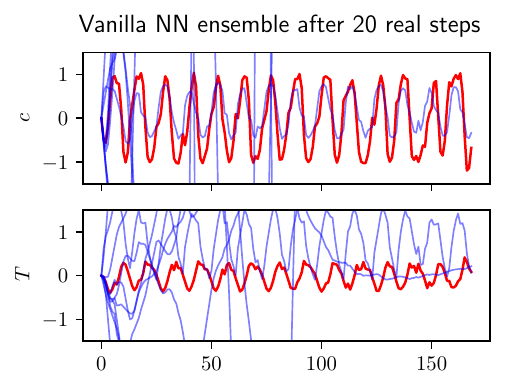} \label{fig:predictions_MLP_20}}
        \subfloat[e][]{\includegraphics[width=0.25\paperwidth]{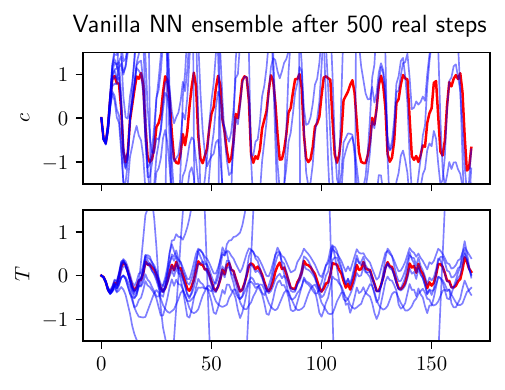} \label{fig:predictions_MLP_500}}
        \subfloat[f][]{\includegraphics[width=0.25\paperwidth]{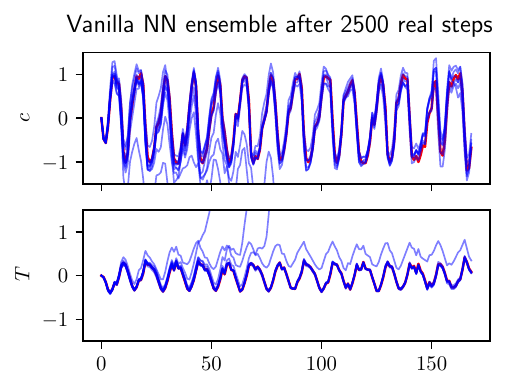} \label{fig:predictions_MLP_2500}}
        \caption{Comparison of closed-loop prediction results of a PINN ensemble and a vanilla NN ensemble after different numbers of data sampling steps in the real environment (Fig. \ref{fig:method_mbpo}, first step). The bounds of each variable (see Tab. \ref{tab:CSTR_bounds_and_ss}) are used for scaling to the [-1, 1] range. The red line depicts the true trajectory of $c$ and $T$. Each blue line corresponds to the closed-loop prediction of one ensemble member. Predictions are chained over the full horizon (168 steps) of the episode.}
        \label{fig:prediction_test}
    \end{figure*}
    The sample efficiency of MBPO is foremost dependent on how many interactions with the real environment are needed until the model ensemble becomes an accurate surrogate of the real environment. Compared to purely data-driven models, PINNs can thrive in settings where few training data are available \cite{Raissi2019Physics-informedEquations}. Here, we aim to confirm that the generally improved performance of the physics-informed variants (\SIKoopPIRLBounds{}, \PIRLMLP{}) compared to their non-physics-informed counterparts (\SIKoopRLBounds{}, \RLMLP{}) is indeed due to the PINNs becoming accurate predictors of the real system behavior more quickly than the vanilla NNs. We randomly pick one of the test trajectories produced by a fully trained \SIKoopPIRLBounds{} controller. Then, we randomly pick one of the \SIKoopPIRLBounds{} and \SIKoopRLBounds{} training runs. We test the ensembles that were produced by the chosen training runs after 20, 500, and 2500 environment steps. Fig. \ref{fig:prediction_test} shows the results of these tests. It is evident from Fig. \ref{fig:predictions_PINN_20} and \ref{fig:predictions_MLP_20} that the data from 20 steps is not sufficient to reliably learn accurate models. However, the predictions of the PINN ensemble diverge less strongly than those of the vanilla NN ensemble. After 500 steps in the real environment (Fig. \ref{fig:predictions_PINN_500} and \ref{fig:predictions_MLP_500}), all but one of the PINNs provide highly accurate predictions, whereas the predictions of the vanilla NN ensemble look comparable to those of the PINN ensemble after 20 steps. After 2500 steps, all members of the PINN ensemble provide accurate predictions over the full 168-step horizon of the test episode. Most members of the vanilla NN ensemble also remain accurate over the full episode; however, some still diverge from the true trajectory after some time. Note that (i) during MBPO policy optimization (see Sec. \ref{sec:results_training_RL}), model predictions are chained only up to eight times and that (ii) for each prediction, a different ensemble member is randomly chosen. Thus, MBPO is relatively robust with respect to compounding model errors. This explains why one can expect to obtain good control performance well before one can expect the ensemble to converge to accurate closed-loop predictions over a long time horizon (cf. Fig. \ref{fig:trainingresults} and Fig. \ref{fig:prediction_test}).

    \begin{figure*}[!htb]
        \centering
        \subfloat{\includegraphics[width=0.3\paperwidth]{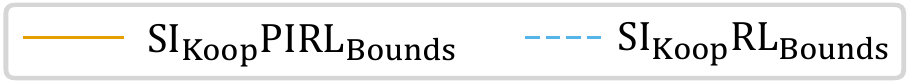} \label{fig:bounds_delta_legend}} \\
        \setcounter{subfigure}{0} 
        \subfloat[a][]{\includegraphics[width=0.25\paperwidth]{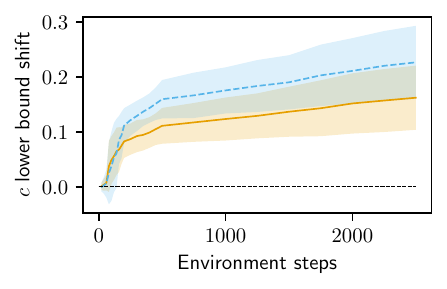} \label{fig:bounds_delta_lb_c}}
        \subfloat[b][]{\includegraphics[width=0.25\paperwidth]{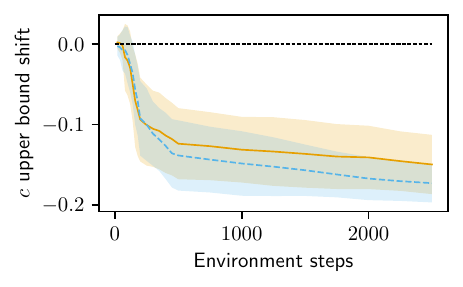} \label{fig:bounds_delta_ub_c}}
        \subfloat[c][]{\includegraphics[width=0.25\paperwidth]{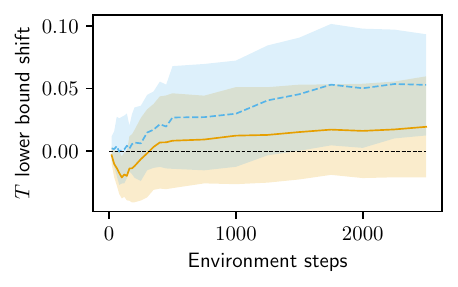} \label{fig:bounds_delta_lb_T}} \\
        \subfloat[d][]{\includegraphics[width=0.25\paperwidth]{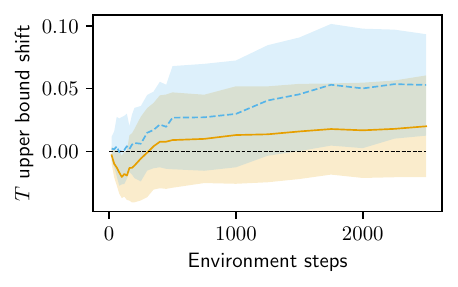} \label{fig:bounds_delta_ub_T}}
        \subfloat[e][]{\includegraphics[width=0.25\paperwidth]{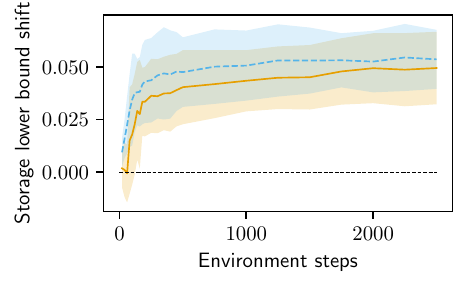} \label{fig:bounds_delta_ub_storage}}
        \subfloat[f][]{\includegraphics[width=0.25\paperwidth]{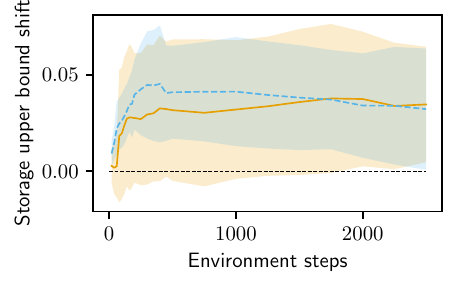} \label{fig:bounds_delta_lb_storage}}
        \caption{Evolution of the parameters $\bm{\theta}_{\text{B}}$ during training. Lines depict the average of the respective parameter over 10 training runs; the shaded regions depict one standard deviation across the different training runs. The adaptation of the bounds is performed with respect to the scaled $c$ and $T$ variables (both variables scaled to the [-1,1] range using their bounds given in Tab. \ref{tab:CSTR_bounds_and_ss}). We leave the product storage in its original [0,6] range.}
        \label{fig:bounds_delta}
    \end{figure*}
    As explained in Sec. \ref{sec:results_training}, in training the \SIKoopPIRLBounds{} and \SIKoopRLBounds{} controllers, $\bm{\theta}_{\text{K}}$ are trained only via SI, whereas $\bm{\theta}_{\text{B}}$ are trained only via PPO using imagined policy rollouts using the learned model ensemble. This means that besides their good performance (see Fig. \ref{fig:trainingresults}), the \SIKoopPIRLBounds{} and \SIKoopRLBounds{} controllers have another valuable property: the (very few) parameters $\bm{\theta}_{\text{B}}$ that are used to optimize the controller for task-optimal control performance are intuitively interpretable. Each parameter in $\bm{\theta}_{\text{B}}$ modifies one of the bounds in the OCPs of the eNMPC (see Eq. \eqref{eq:Koopman_eNMPC}), i.e., the lower and upper bounds of $c$, $T$, and the storage level. Fig. \ref{fig:bounds_delta} shows the evolution of $\bm{\theta}_{\text{B}}$ during the \SIKoopPIRLBounds{} and \SIKoopRLBounds{} training runs. For both variants and across all trained eNMPCs, the bounds of $c$ are tightened (see Fig. \ref{fig:bounds_delta_lb_c} and \ref{fig:bounds_delta_ub_c}), thus decreasing the likelihood of constraint violations. Fig. \ref{fig:bounds_delta_lb_T} and \ref{fig:bounds_delta_ub_T}, which depict the adaptation of the bounds of $T$, are less conclusive. Here, the observed adaptations to the bounds are smaller than those observed for $c$. Furthermore, the directions of the adaptations do not align across all training instances. Since in our controller tests, the bounds of $c$ were violated frequently (see Fig. \ref{fig:control_trajectories}), whereas no violations of the $T$ bounds were observed, it makes intuitive sense that optimizing $\bm{\theta}_{\text{B}}$ leads to a tightening of the $c$ bounds and a less decisive adaptation of the $T$ bounds. Fig. \ref{fig:bounds_delta_lb_storage} shows that a small but consistent back-off from the lower bound of the storage is learned. The adaptation of the upper bound of the storage is less decisive (see Fig. \ref{fig:bounds_delta_ub_storage}). This is not surprising since, like the bounds of $T$, the upper storage bound is not violated by any of the controllers during training.

\section{Conclusion}\label{sec:conclusion}
    We present a method that aims to increase the sample efficiency of RL-based training of data-driven (e)NMPCs for specific control tasks. To that end, we combine our previously published approach (\cite{mayfrank2024KoopmanPPO}) for turning Koopman (e)NMPCs into automatically differentiable policies that can be trained using RL methods with a physics-informed version of the MBPO algorithm (\cite{Janner2019MBPO}), a state-of-the-art Dyna-style model-based RL algorithm. Our method iterates between three steps (see Fig. \ref{fig:method_mbpo}): First, the Koopman (e)NMPC gathers data about the system dynamics by interacting with the physical system. Second, the Koopman model that is used in the (e)NMPC is fitted to the data via SI. Furthermore, an ensemble of NNs is fitted to the data. If (partial) knowledge of the system dynamics is available, physics-informed training can be utilized to increase the accuracy of the NN ensemble. Third, a surrogate environment is constructed using the NN ensemble to simulate the dynamics of the environment. In conjunction with the PPO algorithm (\cite{schulman2017proximal}), this surrogate environment is used to optimize the Koopman (e)NMPC by adapting the variable bounds that are imposed in the OCPs of the (e)NMPC.

    We validate our method using a demand response case study (\cite{mayfrank2024KoopmanPPO}) based on a benchmark CSTR model (\cite{flores2006simultaneous}). The case study involves nonlinear dynamics and hard constraints on system variables; however, due to its small scale it is far less complex than many real-world systems. We compare the performance of our method to that of Koopman eNMPCs trained solely via iterative SI and to NN policies trained via (physics-informed) model-based RL. We find that our method (see Section \ref{sec:method_PIMBRL-Koopman}) outperforms all other tested approaches when applied to our case study: It reaches higher rewards and does so with better sample efficiency and lower variance between differently seeded training instances, resulting in improved economic performance and constraint satisfaction compared to the benchmark methods.

    Although our method for the training of task-optimal Koopman (e)NMPCs achieves excellent sample efficiency in our case study, the training process incurs a high computational cost. This cost is mainly driven by the policy optimization step (see Fig. \ref{fig:method_mbpo}), which involves (i) numerous interactions of the controller with the learned surrogate environment and (ii) differentiating through the OPCs of the (e)NMPC in order to execute policy gradient updates. However, these computationally intensive steps of our approach \textit{do not} need to be executed in real-time. The only step where computations need to be done in real-time is the data sampling step (see Fig. \ref{fig:method_mbpo}). However, while this step involves inference of the current version of the Koopman (e)NMPC, it does not involve backpropagation and parameter updates. The computational burden of merely evaluating a Koopman (e)NMPC is relatively low since it is predominantly driven by the need to solve a convex OCP. Therefore, we assume that our method should be scalable to large systems. Our approach could offer concrete benefits for the control of various real-world systems where mechanistic models cannot be used for predictive control. Future work should, therefore, validate our method on case studies matching the scale and complexity of challenging real-world control problems.

    Recent works aim to improve upon the MBPO algorithm (e.g., \cite{frauenknecht2024trust}) and Dyna-style model-based RL in general (e.g., \cite{frauenknecht2025rollouts}). These methods do not require a specific policy architecture. Therefore, they do not interfere with our approach and could be combined with our method for potentially even better performance. Another avenue of possible future research is combining our method with approaches for learning disturbance estimators for offset-free Koopman MPC (e.g., \cite{SonKwon2021, SonKwonJProc2022}): Instead of learning modifications to the state bounds, a task-optimal disturbance estimator could be learned to estimate the disagreement between the Koopman model and the PINN ensemble.

\section*{Declaration of Competing Interest}
We have no conflict of interest.

\section*{Acknowledgements}
This work was performed as part of the Helmholtz School for Data Science in Life, Earth and Energy (HDS-LEE) and received funding from the Helmholtz Association of German Research Centres.

Part of this work was funded by the Deutsche Forschungsgemeinschaft (DFG, German Research Foundation) – 466656378 – within the Priority Programme “SPP 2331:Machine Learning in Chemical Engineering”.

\section*{Declaration of generative AI and AI-assisted technologies in the writing process}
During the preparation of this work Daniel Mayfrank used Grammarly in order to correct grammar and spelling and to improve style of writing. After using this tool, all authors reviewed and edited the content as needed and take full responsibility for the content of the publication.

\section*{Nomenclature}

\subsection*{Abbreviations}
\noindent\begin{table}[H]
\begin{tabular}{ll}
    NN        & neural network \\
    CSTR      & continuous stirred-tank reactor \\
    DAE       & differential-algebraic equation \\
    (e)(N)MPC & (economic) (nonlinear) model \\
              & predictive control \\
    MBPO      & model-based policy optimization \\
    MDP       & Markov decision process \\
    MLP       & multilayer perceptron \\
    MSE       & mean squared error \\
    OCP       & optimal control problem \\
    PDE       & partial differential equation \\
    PINN      & physics-informed neural network \\
    PPO       & proximal policy optimization \\
    RHS       & right-hand side of equation \\
    RL        & reinforcement learning \\
    SI        & system identification
\end{tabular}
\end{table}

\subsection*{Greek Symbols}
\begin{table}[H]
\begin{tabular}{ll}
    $\alpha$        & reward calculation hyperparameter \\
    $\bm{\theta}$   & learnable parameters of controller \\
    $\lambda$       & PINN loss weight hyperparameter \\
    $\bm{\mu}$      & expected value for action selection \\
    $\bm{\pi}$      & policy \\
    $\rho$          & CSTR production rate \\
    $\bm{\sigma}$   & standard deviation for action selection \\
    $\tau$          & PINN time \\
    $\bm{\phi}$     & learnable parameters of critic \\
    $\Phi$          & MPC stage cost \\
    $\bm{\psi}$     & encoder MLP \\
    $\bm{\omega}$   & learnable parameters of dynamic model \\
\end{tabular}
\end{table}

\subsection*{Latin Symbols}
\begin{table}[H]
\begin{tabular}{ll}
    $\bm{A}$        & autoregressive part of Koopman \\
                    & dynamics matrix \\
    $\bm{B}$        & external input part of Koopman \\
                    & dynamics matrix \\
    $c$             & CSTR product concentration \\
    $\bm{C}$        & decoder matrix of Koopman model \\
    $\mathcal{D}$   & State transition database \\
    $F$             & CSTR coolant flow rate \\
    $\bm{g}$        & inequality constraints \\
    $\bm{h}$        & equality constraints \\
    $l$             & storage filling level \\
    $m$             & control input dimensionality \\
    $M$             & penalty factor for slack variables \\
    $n$             & state dimensionality \\
    $N$             & Koopman state dimensionality \\
    $\bm{\mathcal{N}}$ & normal distribution \\
    $\bm{NN}$       & neural network function representation \\
    $p$             & electricity price \\
    $r$             & reward \\
    $R$             & reaction rate \\
    $\bm{s}$        & slack variables \\
    $t$             & time \\
    $T$             & CSTR temperature \\
    $\mathrm{T}$    & set of discrete time steps \\
    $\bm{u}$        & control variables \\
    $\bm{x}$        & system state variables \\
    $\bm{z}$        & Koopman state variables

\end{tabular}
\end{table}

\subsection*{Subscripts}
\begin{table}[H]
\begin{tabular}{ll}
    $\text{ss}$ & steady-state \\
    $t$         & discrete time step \\
\end{tabular}
\end{table}

\subsection*{Superscripts}
\begin{table}[H]
\begin{tabular}{ll}
    $\Dot{}$    & time derivative \\
    $*$         & indicates optimality \\
    $\hat{}$    & denotes model prediction
\end{tabular}
\end{table}

\section*{CRediT authorship contribution statement}
Daniel Mayfrank: Conceptualization, Methodology, Software, Investigation, Writing - original draft, Writing - review \& editing, Visualization

\noindent Mehmet Velioglu: Conceptualization, Methodology, Software, Investigation, Writing - review \& editing, Writing - original draft

\noindent Alexander Mitsos: Conceptualization, Writing - review \& editing, Supervision, Funding acquisition

\noindent Manuel Dahmen: Conceptualization, Methodology, Writing - review \& editing, Supervision, Funding acquisition

\section*{Author contributions}
\begin{itemize}
    \item DM came up with the initial idea of this work, developed the core method, implemented everything except the PINN models, analyzed and visualized the results, and wrote the original draft of all sections except the PINN-specific parts of Sections \ref{sec:results_casestudy} and \ref{sec:results_training}.
    \item MV implemented the PINN models and wrote the original draft of the PINN-specific parts of Sections \ref{sec:results_casestudy} and \ref{sec:results_training}.
    \item MD and AM contributed to conceptualization of the work and provided supervision. MD also contributed to methodology development. 
    \item All authors have reviewed and edited the manuscript.
\end{itemize}

\appendix

\bibliographystyle{apalike}
  \renewcommand{\refname}{Bibliography}


\end{document}